\documentclass[journal=jacsat,manuscript=article]{achemso}
\usepackage[T1]{fontenc}
\usepackage{amsmath,amssymb,booktabs,longtable,array,graphicx}
\usepackage{xcolor}
\usepackage{xurl}
\usepackage{hyperref}
\hypersetup{colorlinks=true,linkcolor=blue!45!black,urlcolor=blue!45!black,citecolor=blue!45!black}
\newcommand{\wb}{$\omega$B97M-V}
\newcommand{\kcal}{kcal\,mol$^{-1}$}
\newcommand{\Eh}{E_{\mathrm h}}

\newcommand{\NER}{\mathrm{NER}}

\newcommand{\papertitle}{Agent-Guided Discovery of Exchange--Correlation Functionals with Physics-based Validation}
\graphicspath{{Figures/}}
\usepackage{caption}
\usepackage{placeins}
\SectionNumbersOn
\setkeys{acs}{articletitle=true,doi=true}
\title{\papertitle}
\author{Achuth Chandrasekhar}
\affiliation{Mechanical Engineering, Carnegie Mellon University, Pittsburgh, PA 15213, USA}
\author{Amir Barati Farimani}
\email{barati@cmu.edu}
\affiliation{Mechanical Engineering, Carnegie Mellon University, Pittsburgh, PA 15213, USA}
\alsoaffiliation{Biomedical Engineering, Carnegie Mellon University, Pittsburgh, PA 15213, USA}
\alsoaffiliation{Chemical Engineering, Carnegie Mellon University, Pittsburgh, PA 15213, USA}
\alsoaffiliation{Materials Science and Engineering, Carnegie Mellon University, Pittsburgh, PA 15213, USA}
\alsoaffiliation{Machine Learning Department, Carnegie Mellon University, Pittsburgh, PA 15213, USA}

\abbreviations{DFT,XC,MCP,SCF,WRMSD,NER}

\usepackage{listings}
\usepackage[most]{tcolorbox}
\definecolor{taskblue}{HTML}{215A82}
\definecolor{modelgreen}{HTML}{24715D}
\definecolor{feedbackorange}{HTML}{99602B}
\definecolor{proposalviolet}{HTML}{625080}
\lstdefinestyle{transcript}{basicstyle=\ttfamily\small,breaklines=true,breakatwhitespace=false,columns=fullflexible,keepspaces=true,showstringspaces=false,tabsize=2}
\newtcblisting{taskbox}[1]{listing only,breakable,colback=taskblue!4,colframe=taskblue,title={#1},fonttitle=\bfseries,listing options={style=transcript},boxrule=0.5pt}
\newtcblisting{modelbox}[1]{listing only,breakable,colback=modelgreen!4,colframe=modelgreen,title={#1},fonttitle=\bfseries,listing options={style=transcript},boxrule=0.5pt}
\newtcblisting{feedbackbox}[1]{listing only,breakable,colback=feedbackorange!5,colframe=feedbackorange,title={#1},fonttitle=\bfseries,listing options={style=transcript},boxrule=0.5pt}
\newtcblisting{proposalbox}[1]{listing only,breakable,colback=proposalviolet!4,colframe=proposalviolet,title={#1},fonttitle=\bfseries,listing options={style=transcript},boxrule=0.5pt}

\begin{document}
\begin{abstract}
Accurate, compact exchange--correlation functionals are central to predictive density functional theory. Symbolic search and language-model agents can propose new functional forms, but it remains unclear whether their gains survive self-consistency and external evaluation, whether the new terms carry physical content, whether agents outperform evolutionary search, and how language-model backbones compare. We connect frontier LLM agents to a GPU evaluator through the Model Context Protocol, with training-only fitting, persistent memory, and self-consistent and external evaluation. Opus-5.5 agents passed the screening gate with about 40-fold fewer tokens than Codex (GPT-6-Astra) agents and found a candidate with lower validation and held-out errors; only this candidate advanced to self-consistent evaluation. The resulting 13-parameter functional, O159, multiplies same-spin correlation by a smooth density window. Refitted controls favor a window that scales the entire same-spin factor, reducing it by 15--30\% at its peak, and that acts on reaction energies mainly in near-iso-orbital regions; against a refitted no-window control, it lowers heavy-atom total-energy errors on GSCDB137 by 29\%. With about 400 fits each, SyFES evolution did not improve on its GAS22 seed, whereas the agents found forms 0.10--0.13 \kcal\ lower in validation error by adding an explicit density scale. Self-consistent O159 lowers the held-out MGCDB84 test error of $\omega$B97M-V by 11.7\%, and its mean normalized error ratio on 36 GSCDB137 datasets (1.139) is comparable to $\omega$B97M-V (1.144) and below GAS22 (1.190). Frontier-LLM agents thus produced O159, a compact functional whose improvement over $\omega$B97M-V survives self-consistent, physics-based validation.
\end{abstract}

\section{Introduction}
\label{sec:introduction}
Accurate exchange--correlation functionals are essential to the predictive power of Kohn--Sham density functional theory, which makes molecular electronic-structure calculations practical by representing the unknown many-electron contribution through a functional of the density.\cite{kaplan2023predictive, mardirossian2017thirty} Improvements to this approximation can increase the reliability of predictions across molecular chemistry while preserving the computational accessibility that makes density functional theory useful. Compact analytic forms are especially attractive: they are efficient to evaluate, expose their mathematical structure to physical analysis, and can be incorporated directly into self-consistent electronic-structure calculations.\cite{ma2022evolving} Discovering such forms with improved accuracy across chemical environments is therefore an important objective for automated scientific discovery.

Functional development increasingly combines physical insight and reference data with automated exploration.\cite{kaplan2023predictive, pederson2022machine} Recent data-driven approaches have learned exchange--correlation models from molecular energies and densities,\cite{nagai2020completing, dick2020machine} while differentiable Kohn--Sham training incorporates self-consistency into the learning loop.\cite{kasim2021learning, li2021kohn} Modern range-separated hybrids such as \wb\ integrate semilocal descriptors, exact exchange, and nonlocal correlation in compact expressions selected and fitted against diverse reference data.\cite{mardirossian2017thirty, head2016v} Symbolic Functional Evolutionary Search (SyFES) broadened this strategy by evolving computational instructions to explore the functional form itself. Starting from \wb, it produced GAS22, whose improvements identified on fixed reference densities persisted in subsequent self-consistent-field (SCF) calculations.\cite{ma2022evolving} More recently, SAFS26 used language-model agents to plan modifications, implement functional expressions, and interpret evaluation feedback through a structured evolutionary archive.\cite{duston2026agentic} In a nonempirical setting, an LLM agent paired with an LLM critic and deterministic exact-constraint verification revised r2SCAN toward larger band gaps without fitting any bonded system, locating stronger candidates than a uniform random-search control.\cite{adhikari2026agentic} The wider strategy of generating executable
proposals and ranking them with numerical feedback has also enabled language-model-guided
program search in mathematics.\cite{romera2024mathematical} These advances establish a foundation for agents that use numerical feedback and accumulated search experience to guide the discovery of explicit mathematical forms.

For empirically fitted functionals, it has not yet been established whether agent-discovered expressions retain their search-stage gains once they are physically screened, made self-consistent, and evaluated beyond the selection benchmark. Fixed-density evaluation enables rapid exploration, whereas self-consistent calculations test the orbitals and density generated by the candidate's own potential.\cite{sim2022improving} Physical and numerical screening guides the search toward suitable expressions, while held-out and external evaluations test their predictive value.\cite{kaplan2023predictive, furness2020accurate, kapoor2023leakage} GSCDB137 offers a broader assessment through transition-metal chemistry, electric-field responses, vibrational properties, heavy elements, and revised reference values.\cite{liang2025gold} Its datasets without MGCDB84 antecedents (hereafter, new datasets) provide a particularly useful opportunity to examine performance beyond the benchmark used for discovery. Bringing these stages together with persistent search memory allows frontier LLM agents to be assessed by the functional forms they produce and the subsequent calculations those forms support.

Two further gaps limit what such searches currently offer chemistry. First, joint refitting distributes credit across every coefficient, so the physical role of a newly introduced term cannot be read from its expression. It must be established with refitted controls, localized in descriptor space, and checked against data not used for selection; otherwise a discovered functional remains an accurate fit without a transferable design principle. Second, the r2SCAN revision compared its agent with uniform random sampling of a fixed term catalog,\cite{adhikari2026agentic} but, to our knowledge, language-model-guided searches have not been compared with established non-LLM symbolic evolution at a matched number of trained forms on identical data, and different language-model backbones have not been compared on a common evaluator.

\begin{figure}[!htbp]
\centering
\includegraphics[width=\linewidth]{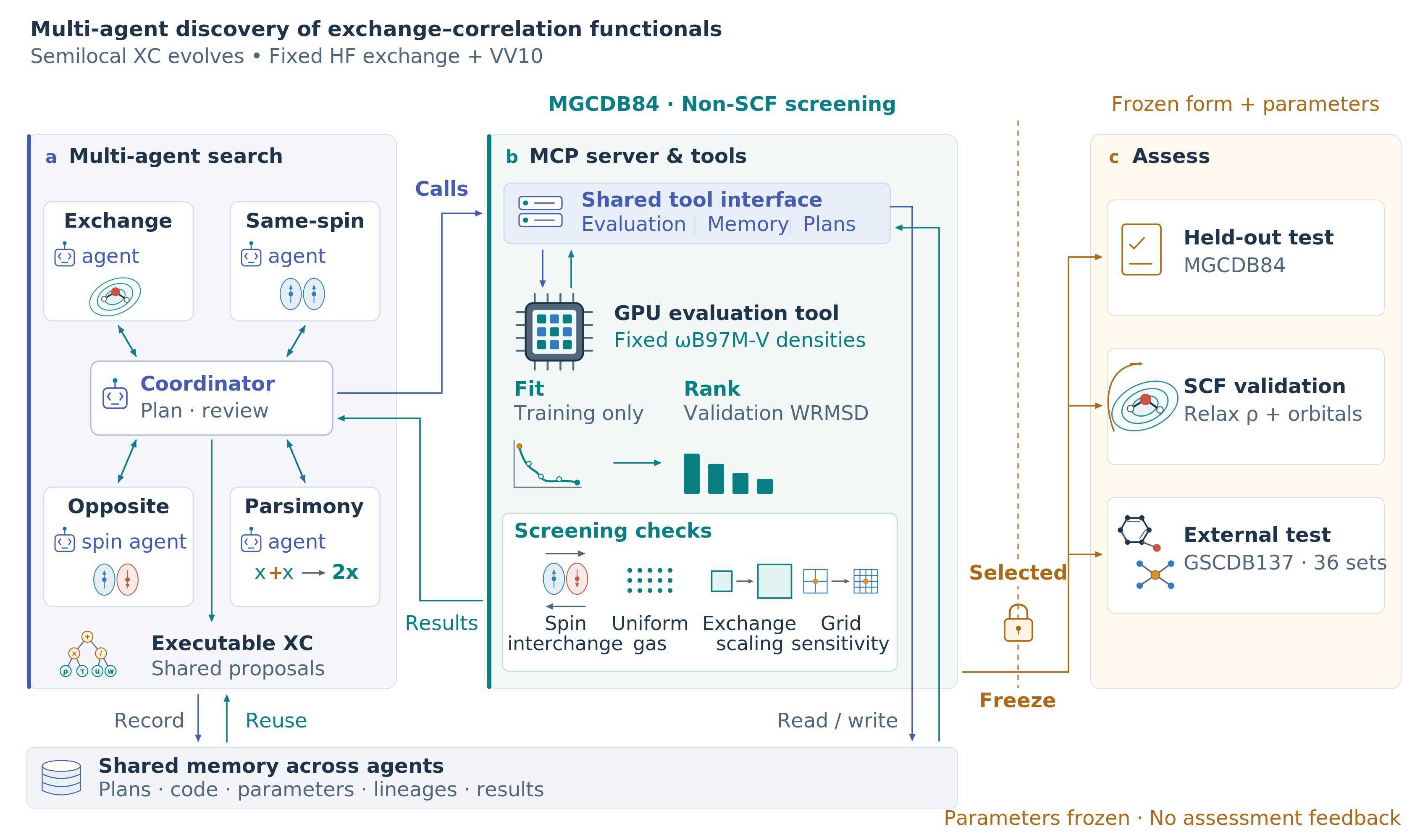}
\caption{Workflow for agent-guided exchange--correlation functional discovery and evaluation. (a) A coordinator and thematic agent islands generate executable proposals, drawing on shared plans, lineages, results, and summaries. (b) The MCP server connects agents to the GPU evaluator, which fits coefficients on training data, measures validation WRMSD, and applies four screening checks: spin interchange, the uniform-gas exchange limit, exchange-factor scaling, and grid sensitivity. (c) Selected candidates and fitted parameters are frozen before held-out MGCDB84 testing, selected self-consistent-field validation, and external evaluation on the 36-dataset computational subset of GSCDB137. These downstream assessments are separate from discovery; their results were not used to fit, rank, or select candidates.}\label{fig:workflow}
\label{fig:agent-workflow}
\end{figure}

Here we address these gaps by connecting scientifically capable frontier LLM agents to a shared GPU evaluator through the Model Context Protocol (Figure~\ref{fig:workflow}).\cite{hou2026model} LLM agents have automated scientific tasks such as heterogeneous catalyst screening, polymer and peptide design, literature mining, molecular dynamics of polymers and proteins and many more.\cite{chandrasekhar2026catalyst, nigam2026polymer, chen2026pepti, chandrasekhar2026material, zhao2026polyjarvis, pak2026agentic, 10.1039/d6dd00535g, ock2026large, zeng2025llm} Agents propose symbolic expressions and build on recorded plans, parent lineages, and numerical feedback; the evaluator fits coefficients only on training data, scores validation performance, and applies four physical and numerical screening tests. This separation combines flexible exploration with a consistent evaluation procedure, while persistent records preserve the reasoning and computational provenance of the search. The principal result is O159, a 13-parameter functional that introduces a smooth density-window multiplier into same-spin correlation at intermediate density. O159 satisfies all four screening criteria: spin symmetry, the uniform-gas exchange limit, coordinate scaling of the exchange factor, and integration-grid sensitivity. We then use refitted controls that the agents recorded during the search, a descriptor-resolved attribution of the window energy, and held-out MGCDB84 data to determine what this term encodes. An external comparison with the same control on GSCDB137 then tests how far the window's effect transfers. Finally, we compare two language-model backbones, Opus-5.5 and Codex (GPT-6-Astra), on the same evaluator and data, and compare both agent searches with SyFES regularized evolution given a comparable number of fits (about 400).

Refitted controls distill the window into an empirical design rule: it should reduce the magnitude of the entire same-spin factor at intermediate spin density, with no opposite-spin counterpart, $\tau_W/\tau$ gate, or shift to lower density. On external data, its largest benefit over the refitted control is a 29\% lower error for heavy-atom total energies. Under the same evaluator and data, the Opus-5.5 agents reached the SCF gate with about 40-fold fewer tokens than the Codex agents and found a candidate with lower validation and held-out test errors; only this candidate, O159, was advanced to self-consistent evaluation. At a matched number of fits, the SyFES baseline did not improve on its GAS22 seed, whereas the agents found screened forms with lower validation error, all of which add an explicit density scale absent from the SyFES representation. O159 SCF calculations converge for all 5,930 MGCDB84 systems, reduce validation and test WRMSDs relative to \wb\ by 5.7\% and 11.7\%, and give a mean NER on 36 GSCDB137 datasets comparable to that of the local \wb\ reference. The study thus links agent-guided symbolic discovery to a compact, self-consistently validated functional with an interpretable new term, and compares the performance of different LLM backbones.

\section{Methods}
\subsection{Agentic search and numerical evaluator}
The framework exposes functional evaluation, lineage retrieval, memory, and recorded summaries through the Model Context Protocol.\cite{hou2026model} The search archive stores candidate source, fitted parameters, parent identifiers, plans submitted before evaluation, summaries submitted afterward, constraint results, and usage counters. The multi-agent organization assigns thematic exploration to islands and coordination to a lead agent and shared records carry information across separate contexts. The completed Codex run additionally used independent audits every 50 successful evaluations and a queue limited to three concurrent GPU jobs. Its final island allocations were 100 exchange, 120 same-spin, 100 opposite-spin, and 80 parsimony evaluations, after a documented budget transfer. Supporting Information Section~S7 gives the representative prompt, ranking rule, and reconstructed iteration exchanges.

\begin{figure}[!htbp]
\centering
\includegraphics[width=\linewidth]{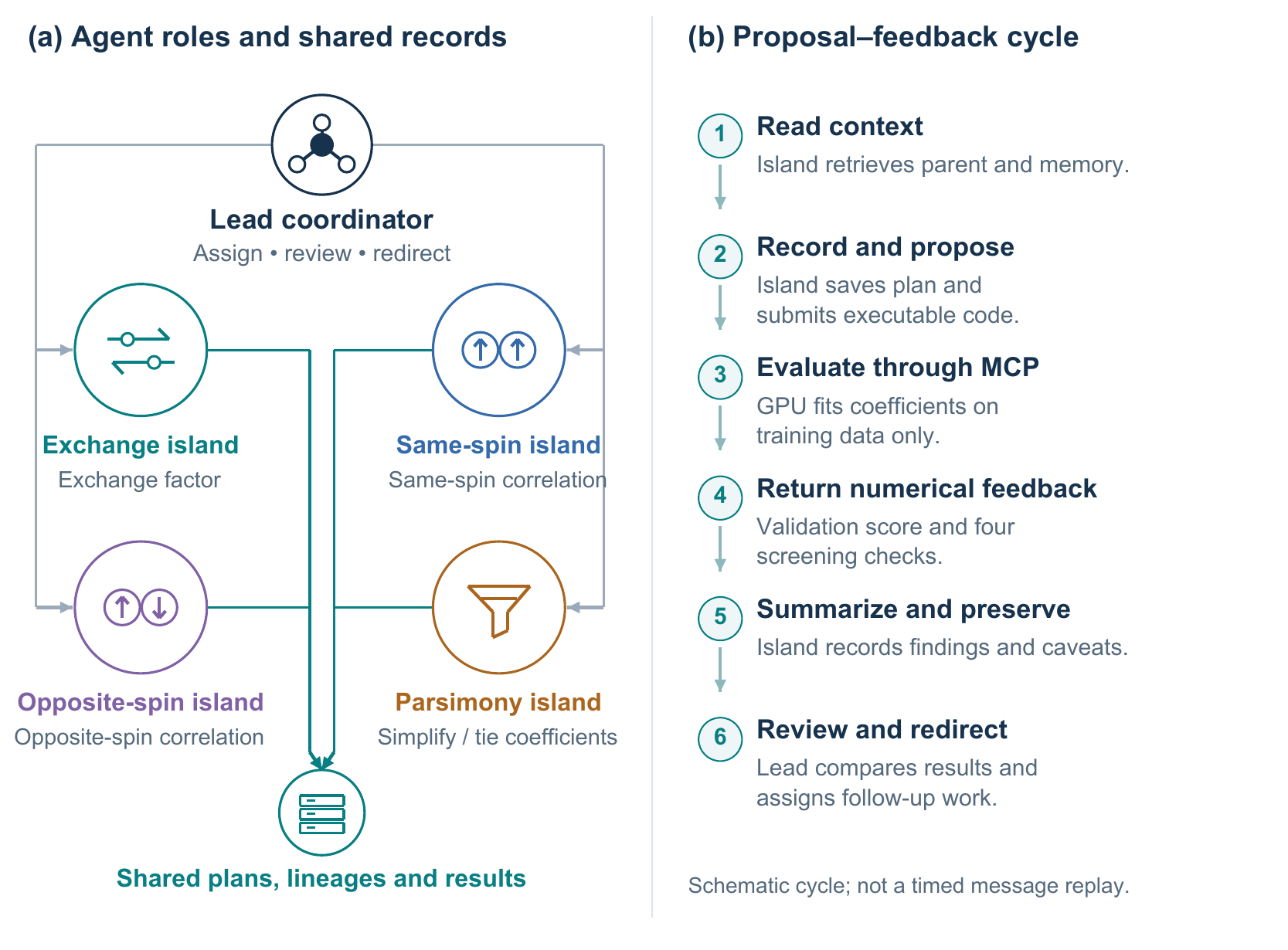}
\caption{Agent responsibilities and communication during symbolic functional discovery. (a) The lead coordinator organizes thematic exchange, same-spin, opposite-spin, and parsimony roles. Shared records preserve plans, source code, fitted parameters, lineages, results, and summaries across agent contexts. (b) A schematic proposal--feedback cycle: an island retrieves context, records a plan and executable proposal, and calls the evaluator through MCP. The evaluator fits coefficients on training data, returns validation scores and four screening outcomes, and the island records a summary for subsequent review and proposals. The schematic shows agent roles and information flow through the proposal--evaluation cycle; held-out and external assessments remain outside this discovery loop.}
\label{fig:agent-roles}
\end{figure}

Figure~\ref{fig:agent-roles} details the division of work and the information exchanged during discovery. Exchange, same-spin, and opposite-spin islands propose changes to their respective semilocal factors, while the parsimony island simplifies expressions or ties coefficients. The coordinator assigns themes, reviews numerical feedback, and redirects follow-up work. Islands communicate through persistent plans, parent lineages, results, and summaries; MCP connects the agents to these shared records and to the numerical evaluator.

The retained Opus-5.5 archive contains 397 non-seed submissions, and the completed Codex (GPT-6-Astra) archive contains 400 successful evaluations plus one failed attempt. These counts include repeated fits and initialization controls, not only distinct symbolic forms. A superseded 200-attempt Codex experiment is documented separately in Supporting Information Section~S6. Opus and Codex denote run archives for the respective model families.

Each proposal is implemented as one functional class. All retained empirical coefficients and introduced empirical shape parameters are fitted on the training objective using automatic differentiation and L-BFGS.\cite{paszke2019pytorch, kochenderfer2019algorithms} Restarts are selected by training loss. The prompt limits a request to three restarts and 500 optimizer iterations; the service imposes a 560 s tool timeout and a usual 410 s fitting budget. Training uses compiled float32 kernels, while final reductions and reported metrics use float64. Three NVIDIA RTX 6000 Ada GPUs provide the worker pool. The recorded environment includes PySCF 2.14.0 and GPU4PySCF 1.8.1.\cite{sun2020recent, li2025introducing, wu2025enhancing} Parameter files, numerical guards, software snapshots, and complete computational settings are supplied in Supporting Information Sections~S1 and S2 and the reproducibility archive.

\subsection{Functional framework, data, and screening}
Candidates modify semilocal exchange and same-spin/opposite-spin correlation within the fixed energy framework
\begin{equation}
E_{xc}=E_{x,\mathrm{sl}}^{\mathrm{SR}}+0.15E_{x,\mathrm{HF}}^{\mathrm{SR}}
+E_{x,\mathrm{HF}}^{\mathrm{LR}}+E_{c,ss}+E_{c,os}+E_c^{\mathrm{VV10}}.
\label{eq:main-frame}
\end{equation}
The range-separation parameter is $\omega=0.3\,a_0^{-1}$, and VV10 parameters are $b=6$ and $C=0.01$.\cite{head2016v, vydrov2010nonlocal} Semilocal factors depend on spin densities, density gradients, and positive kinetic-energy densities. Correlation uses the Stoll partition of modified PW92 local correlation.\cite{perdew1992accurate, stoll1978calculation} A nonhybrid B97M-V seed is retained as a separately labeled reference refit.\cite{mardirossian2015mapping}

MGCDB84 reaction and geometry files were obtained through ACCDB with the released SyFES partition and weighting algorithm.\cite{mardirossian2017thirty, ma2022evolving, morgante2019accdb, SyFEScode2022} After omission of XB51\_15, the implemented split contains 870 training, 2,963 validation, and 1,152 test reactions requiring 5,930 unique molecular calculations. Parameters are fitted on training data; validation guides expression selection. The weighted root-mean-square deviation (WRMSD) is
\begin{equation}
J_S=\left[\frac1{N_S}\sum_{i\in S}w_i
(\Delta E_i-\Delta E_i^{\mathrm{ref}})^2\right]^{1/2},
\label{eq:main-wrmsd}
\end{equation}
with energies in \kcal, reaction-specific weights $w_i$, and reaction count $N_S$. Validation weights include the prescribed factor two. Subsets are assigned as groups, so the partitions have different category compositions. They are not random draws from a common reaction distribution.

Screening uses fixed \wb/def2-QZVPPD reference densities.\cite{head2016v, weigend2005balanced, rappoport2010property} The def2 basis definitions are also catalogued in the Basis Set Exchange.\cite{pritchard2019new} Reference orbitals were converged with density fitting\cite{wu2025enhancing} and then evaluated with exact Coulomb/exchange integrals. Candidate non-SCF energies replace the semilocal contribution at the fixed density while retaining the common exact-exchange and VV10 terms. The reference-density campaign converged all 5,930 systems and consumed 84.6 GPU-h. Cached search metrics and later full-grid non-SCF reevaluations are explicitly distinguished: float32 feature storage introduces small differences even when the final arithmetic uses float64.

The evaluator checks spin interchange on triplet oxygen, the uniform-gas exchange factor at four densities, exchange-factor invariance under coordinate scaling on water, and AE18 atomic integration sensitivity between two grids. The respective tolerances are $10^{-5}\Eh$, absolute $|g_x-0.85|<0.005$, $10^{-4}\Eh$, and 0.015 \kcal; nonfinite values are rejected. The scaling check is applied to the exchange enhancement factor without range attenuation. The search score is proportional to inverse validation WRMSD with a factor $0.9$ per violation. The standard SCF gate requires zero violations and validation below 97\% of the cached published-parameter baseline. The checks target exact conditions and numerical requirements relevant to self-consistent use\cite{kaplan2023predictive, furness2020accurate, bartok2019regularized} and complement empirical accuracy on molecular benchmarks.\cite{pederson2023difference}

\subsection{Frozen evaluation and external scope}
Test evaluation uses the stored fitted coefficients without refitting. For SCF, automatic differentiation of the energy density supplies density, gradient, and kinetic-energy-density potential components to GPU4PySCF.\cite{paszke2019pytorch, li2025introducing, wu2025enhancing} The candidate orbital calculation is initialized from the reference solution and followed by exact-integral energy evaluation. Only O159 and published-parameter GAS22 received complete candidate MGCDB84 SCF campaigns. Candidate energies and potentials come from the PyTorch implementation, whereas the \wb\ reference uses libxc.

To limit computational cost, external evaluation used the 36 GSCDB137 datasets whose largest molecule has at most 500 basis functions: 2,734 entries and 3,487 molecular/field calculations. Sixteen datasets have MGCDB84 antecedents and 20 do not; results are reported for both scopes. The benchmark's input-specific basis sets, grids, effective core potentials, fields, and state-search requirements were retained, with documented CPU fallbacks for affected high-angular-momentum calculations. No GSCDB137 data were used to fit or select candidates.

GSCDB137 performance is the unweighted mean of dataset normalized error ratios (NERs).\cite{liang2025gold} Each dataset's specified error measure is divided by its fixed published normalization; NER is dimensionless and differs from the reaction-weighted MGCDB84 objective. The 36-set evaluation includes AE11 after CPU completion of the Rn calculation, and a 35-set analysis without AE11 quantifies sensitivity to that dataset. Paired signed-rank tests and bootstrap intervals use datasets as sampling units.\cite{gibbons2025nonparametric, holmes2019modern} Numerical convergence qualifications are provided in Supporting Information Section~S4.

\subsection{Window controls, energy attribution, and non-LLM baseline}
All window controls are joint training fits that the agents submitted during the search: variants of the O145/O159 architecture in which the window was removed, moved, reshaped, transferred to another channel, or applied to another base functional. The principal control is O178, the O145 form without the window (11 parameters). The analysis adds fixed-density post-processing of archived parameters, a non-SCF evaluation of O178 on GSCDB137 (about 0.5 GPU-h), and repeated O159 refits (Supporting Information Sections~S2 and S3).

The window energy $E_{\rm win}=E_c[\mathrm{O159}]-E_c[\mathrm{O159};\beta_0=\beta_2=0]$ is attributed to regions of $r_{s\sigma}$ and $w_\sigma$ by an exact decomposition of the change in squared WRMSD, defined in Supporting Information Section~S3.

To test the window against the refitted control on external data, O178 was evaluated non-self-consistently on the 36 GSCDB137 datasets at the stored \wb\ densities used for O159. Because the two functionals share exact exchange and VV10, the O178 energy follows from the archived O159 non-SCF energy plus the difference of their semilocal energies, integrated on the production grids; this agrees with direct O178 evaluation and reproduces the archived O159 results (Supporting Information Section~S3). Before the O178 energies were computed, we recorded a predicted direction (improve, worsen, or neutral) for each dataset on which MGCDB84 training, validation, or test trends bear, with neutral defined as a relative change of the dataset error measure below 1\%. The 13 new datasets for which the pattern offered no basis (transition-metal, field-response, and vibrational sets, and six thermochemical, noncovalent, or barrier-height sets) received no prediction and were not scored. The predictions were recorded after the O159 and \wb\ results on GSCDB137 were known.

As a non-LLM baseline, we ran SyFES regularized evolution on the same split, weights, reference densities, and non-SCF energy expression. The implementation follows the published instruction set, mutation rules, functional-equivalence fingerprints, and evolution settings: population 100, tournament size 25, mutation probability 0.9, and covariance matrix adaptation evolution strategy (CMA-ES) fitting on training data with five restarts, where each new form's fitness is the validation WRMSD of its best-training restart. The evolution was seeded with GAS22 at its published parameters and stopped after 400 trained forms, approximately the number of fits in each retained agent archive. The best forms were re-evaluated in float64 and screened with the same four tests. Implementation details and deviations are given in Supporting Information Section~S5.

\section{Results and Discussion}
\subsection{Compact symbolic changes and validation progress}
\begin{figure}[htbp]\centering
\includegraphics[width=\linewidth]{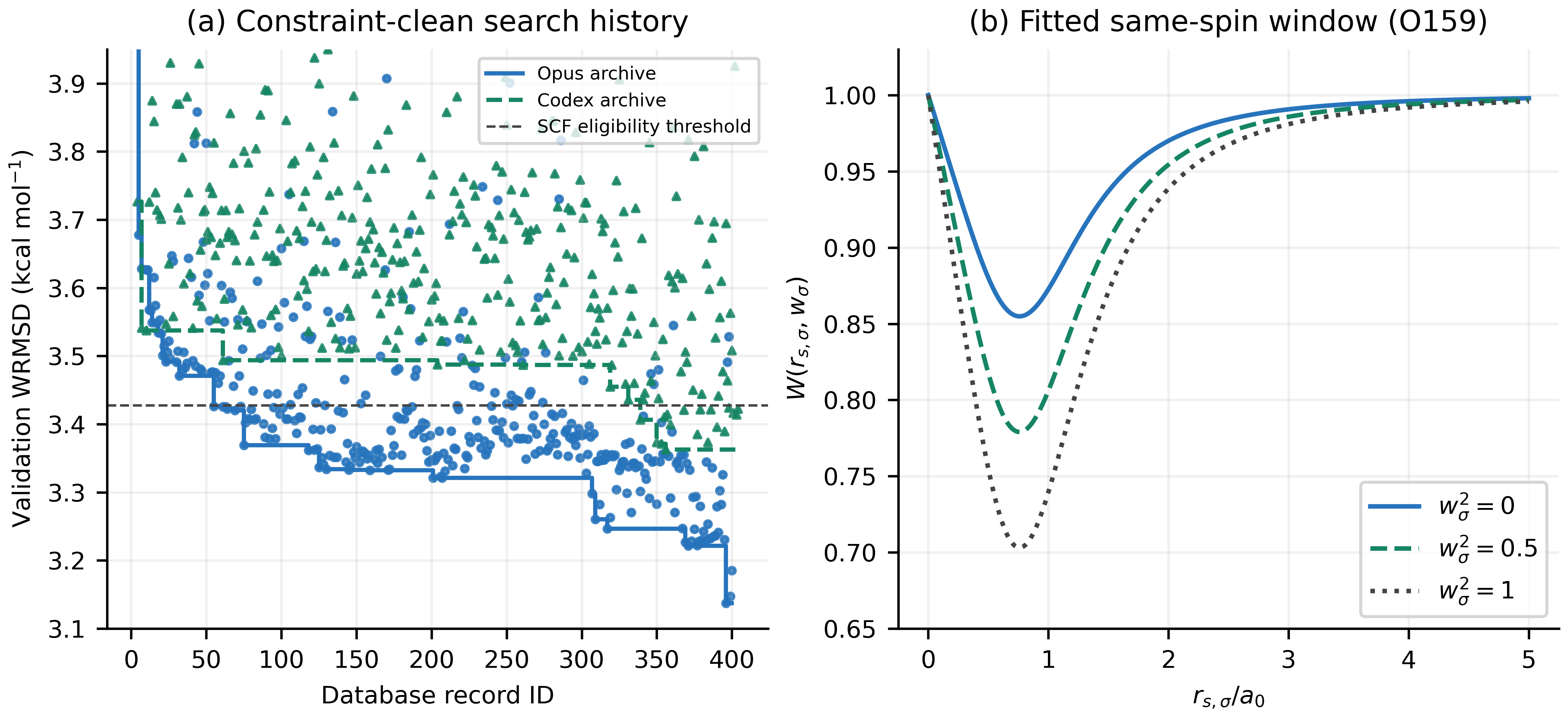}
\caption{Search-stage validation and the O159 density window. (a) Constraint-clean candidate scores and running minima from the retained Opus and Codex archives, indexed by database record ID. The displayed range emphasizes the competitive region; the archive retains all values and audit exclusions. (b) Analytic O159 multiplier at fixed kinetic descriptor, calculated from the evaluated coefficients. The radius refers to one spin density.}\label{fig:search}
\end{figure}

Both retained searches produced constraint-clean candidates below the SCF gate (Figure~\ref{fig:search}). O159 reached cached validation WRMSD 3.332 \kcal\ with 13 parameters. The later O396 candidate reached 3.138 with ten parameters, and C356 reached 3.363 with 14 parameters.

O159's characteristic modification is a multiplier on a GAS22-type same-spin polynomial,
\begin{equation}
g_{ss,\sigma}=P_s(w_\sigma,u_\sigma)
\left[1+(\beta_0+\beta_2w_\sigma^2)
\frac{r_{s\sigma}}{1+r_{s\sigma}^4}\right],
\label{eq:main-window}
\end{equation}
where $r_{s\sigma}=(3/4\pi\rho_\sigma)^{1/3}$ is expressed in bohr, $w_\sigma=(t_\sigma-1)/(t_\sigma+1)$, and $t_\sigma=\tau_{U\sigma}/\tau_\sigma$. The fitted values are $\beta_0=-0.25463509$ and $\beta_2=-0.26628913$ to eight decimals. The exchange and opposite-spin polynomial forms derive from \wb, with all retained coefficients jointly refitted and the opposite-spin constant fixed at one. Here $P_s$ is the GAS22-type same-spin power series in $w_\sigma$ and the bounded reduced gradient $u_\sigma$, and vanishes for the uniform gas. We write $h_p(r)=r/(1+r^p)$, so that O159 uses $h_4$ and its parent O145 uses $h_3$, and $\tau_{W\sigma}=|\nabla\rho_\sigma|^2/(8\rho_\sigma)$ for the von Weizs\"acker kinetic-energy density, and $\alpha_\sigma=(\tau_\sigma-\tau_{W\sigma})/\tau_{U\sigma}$, with $\tau_{U\sigma}$ the uniform-gas value. Supporting Information Section~S2 gives the complete equations and evaluated parameters.

The window approaches unity at high and low density and changes the same-spin factor only at intermediate density; its role is established by refitted controls rather than by the expression alone. Changing the cubic denominator of the parent O145 to the quartic denominator of O159 alters validation by only 0.0006 \kcal. The controls show that this exponent is immaterial, whereas the channel and scope of the window are not. Supporting Information Section~S3 uses these controls to derive an empirical design rule, in which the window reduces the entire same-spin factor by 15--30\% at intermediate spin density with no opposite-spin counterpart, $\tau_W/\tau$ gate, or shift to lower density, and localizes its effect on reaction energies in low-gradient, near-iso-orbital regions.

Later candidates explore different correlations between descriptors. O396 adds same-spin and opposite-spin shell-localized terms to a compressed window form. Its development included a residual scan over training and validation data, so its validation error is not independent of its design. C356 combines a same-spin orbital window with a density-weighted opposite-spin kinetic descriptor, a screened kinetic term, and a coefficient tie that reduces parameter count.

\subsection{Comparison of language-model backbones}\label{sec:backbones}
The two backbones searched the same functional frame with the same evaluator, data split, screening tests, and SCF gate, so their archives are compared under identical evaluation conditions. The Opus-5.5 agents first crossed the SCF gate with a constraint-clean candidate after about 10.2 million assigned tokens (O55), whereas the Codex (GPT-6-Astra) agents first crossed it after 399.9 million (C339). The Opus campaign also reached its principal candidate at about one ninth of the token cost: O159, reached after 49.9 million tokens, has a validation WRMSD of 3.332 \kcal\ with 13 parameters, whereas the best Codex candidate, C356, reached 3.363 \kcal\ with 14 parameters after 440.6 million tokens. O159 is also more accurate on held-out data, where the non-SCF test WRMSDs are 3.818 \kcal\ for O159 and 3.897 \kcal\ for C356. Because C356 did not improve on O159 in non-SCF validation or test error, O159 remained the only agent-discovered candidate advanced to full SCF evaluation on MGCDB84 and GSCDB137. Supporting Information Section~S6 gives the complete token accounting of both runs (251 and 521 million assigned tokens, more than 97\% of them cache reads). Supporting Information Section~S7 illustrates with representative exchanges how the persistent plans, summaries, and audits of both runs document search decisions.

\subsection{Agent-guided and evolutionary symbolic search at a matched number of fits}\label{sec:syfes-baseline}
The SyFES baseline trained 400 distinct forms in 20.5 h of wall time on three GPUs, using 61.1 GPU-h of CMA-ES fitting (mean 550 GPU-s and $1.2\times10^5$ CMA-ES evaluations per form). Of these forms, 368 gave finite losses. A further 1,079 children were functionally equivalent to earlier forms and inherited their cached fitness without training. None of the trained forms improved on the validation WRMSD of the published GAS22 seed, 3.462 \kcal\ (Table~\ref{tab:syfes}). The best new form, the 95th trained (population identifier 174), adds a $u^2$ term with unit coefficient to the same-spin factor of GAS22 and refits all 14 parameters. It passes all four screening tests and removes GAS22's uniform-gas exchange violation, but its training and validation errors (3.444 and 3.506 \kcal) are higher than those of the seed and of the L-BFGS training refits of the unmodified GAS22 form (Codex: 3.266 and 3.489 \kcal); the added term brings no validation gain. The next nine forms fit training data as well as or better than GAS22 (3.188--3.269 \kcal) yet validate worse (3.554--3.595 \kcal), and all fail the uniform-gas exchange test.

\begin{table}[htbp]
\centering\small
\setlength{\tabcolsep}{3.5pt}
\caption{Agent-guided and non-LLM symbolic search at a matched number of trained forms on the MGCDB84 SyFES split (non-SCF WRMSD, \kcal).}\label{tab:syfes}
\begin{tabular}{lrllrrrr}
\toprule
Search & Fits & Seed (val.) & Best new form & Training & Validation & GPU-h & Tokens (M) \\
\midrule
SyFES evolution & 400 & GAS22 (3.462) & SyFES-174 & 3.444 & 3.506 & 61.1 & 0 \\
Opus agents & 397 & refits$^a$ & O159 & 3.252 & 3.332 & 11.1 & 251 \\
 & & & O396$^b$ & 3.140 & 3.138 & & \\
Codex agents & 400$^c$ & refits$^a$ & C356 & 3.207 & 3.363 & 8.6 & 521 \\
\bottomrule
\end{tabular}
\par\smallskip\noindent\begin{minipage}{0.97\linewidth}\footnotesize\raggedright All best new forms pass the four screening tests; the published GAS22 seed fails the uniform-gas exchange test. SyFES records are distinct trained forms; agent records include repeated fits and initialization controls. SyFES values are float64 re-evaluations of stored parameters; agent values are cached search metrics. GPU-h is CMA-ES training time for SyFES and summed evaluator time for the agents; during the SyFES run the GPUs were shared with another service. O159-form refits from other initializations reach 3.380 \kcal\ (O195). $^a$Training refits of \wb\ (validation 3.870 Opus, 3.789 Codex), GAS22 (3.490 and 3.489; one violation), and nonhybrid B97M-V. $^b$Best record of the second Opus campaign (IDs 201--400); its development used a training-plus-validation residual scan. $^c$Plus one failed attempt.\end{minipage}
\end{table}

With a similar number of training fits, the agent searches found screened forms 0.13 (O159) and 0.10 \kcal\ (C356) below the published GAS22 validation error, and 0.16 and 0.13 \kcal\ below their own GAS22 training refits, whereas evolution from GAS22 found none. O396 is 0.32 \kcal\ lower. The O159 and C356 margins are comparable to the spread of repeated fits of one form (up to 0.09 \kcal; Supporting Information Section~S3), and the agent archives took their minimum on validation over repeated fits and controls, whereas SyFES never refits an equivalent form. The archive shows where the margin comes from. O178, the O145 frame without its window, uses only the scale-invariant variables available to SyFES and validates at 3.481 \kcal, no better than GAS22 (3.462 published, 3.490 refitted). Every best agent form uses an explicit density scale ($r_{s\sigma}$ in O159 and O396, $r_s$ in C356), and the best Codex record without one (C398, 3.424 \kcal) relies on spin-polarization weighting, also outside the SyFES features, and improves on GAS22 by only 0.04 \kcal. Within the representation shared with SyFES, neither search improved on GAS22 at this budget. The agents' advantage therefore came from importing an ingredient outside that representation, one suggested by their prompt and present in the SAFS26-b functional on their reference sheet, and testing it with their own controls. Fitting time and language-model cost point in opposite directions: the agent archives used 11.1 and 8.6 GPU-h of warm-started L-BFGS fitting and evaluation but 251 and 521 million assigned tokens (Section~\ref{sec:backbones}), whereas SyFES used 61.1 GPU-h of gradient-free CMA-ES from random starts and no language model.

\begin{table}[htbp]
\centering\small
\caption{Differences between the two search workflows that bear on the comparison in Table~\ref{tab:syfes}.}\label{tab:syfes-factors}
\begin{tabular}{p{0.72\linewidth}l}
\toprule
Factor & Favors \\
\midrule
Explicit density variable $\rho_\sigma$ (or $r_{s\sigma}$) available; SyFES instructions act only on the scale-invariant $x^2=X_\sigma$ and $w$, from which $\tau_{W\sigma}/\tau_\sigma$ and $\alpha_\sigma$ follow & agents \\
Prompt listed $r_s$-dependent correlation and $\tau_W/\tau$-based self-correlation corrections among example directions; reference sheet listed prior forms, including the multiplicative same- and opposite-spin $r_s$ corrections of SAFS26-b; second Opus campaign given window-related island themes & agents \\
Prompts quoted published validation and test errors of reference functionals and set beating GAS22 and SAFS26-a as a goal; no candidate test value was available during the first Opus campaign & agents \\
Category-resolved validation errors instead of a scalar fitness; warm starts from parent coefficients; screening penalty in the score & agents \\
Minimum taken on validation over repeated fits of the same form & agents \\
Seed is the validation minimum of the original 29,628-form search and keeps its published parameters, whereas children receive only a training-set CMA-ES fit & SyFES seed \\
Gradient-free CMA-ES from random starts; one evolution with three workers; PBEC instruction applied to $x^2$ and $w$ only & neither (not tested) \\
\bottomrule
\end{tabular}
\end{table}

These differences are summarized in Table~\ref{tab:syfes-factors}. CMA-ES applied to the GAS22 form approaches a lower training loss than the published parameters (about 3.205 versus 3.270 \kcal, still decreasing) in a different region of parameter space, and the near-miss forms, which fit training better and validate worse, are consistent with this fitting asymmetry. During the second Opus campaign, the SCF test errors of O159 and GAS22, computed after the first campaign for reporting, were present in the shared notes; the agents were instructed not to use them for parent choice, search direction, or ranking. The Codex run was instructed to work without access to any earlier search results. The comparison therefore tests two complete workflows, including their search spaces and guidance.

The validation advantage did not transfer to held-out data. No new SyFES form was evaluated on the test split, so the relevant held-out comparison is with GAS22 itself, the product of an original search that explored about 20 times as many forms (29,628 versus about 1,480 proposed children here). GAS22 (test 3.599 \kcal, non-SCF) is more accurate than O159 (3.818), O396 (3.841), and C356 (3.897), and all three also trail a training refit of GAS22 (3.628). What distinguishes the agent records is the experiments they contain: hypothesis-directed controls, such as moving the window to another channel or gating it by $\tau_{W\sigma}/\tau_\sigma$, that the randomly sampled SyFES mutations do not target and its scalar fitness does not record. These controls supply the window analysis of Supporting Information Section~S3.

\subsection{Self-consistency preserves the O159 improvement}
Table~\ref{tab:main-comparison} brings together the primary MGCDB84 split and the completed GSCDB137 validation subset. On MGCDB84, O159 SCF lowers validation WRMSD from the reference's 3.533 to 3.330 \kcal\ and test WRMSD from 4.309 to 3.805, reductions of 5.7\% and 11.7\%. Its non-SCF and SCF errors differ by at most 0.013 \kcal, so O159 retains its validation and test improvements over \wb\ after self-consistency. GAS22 nevertheless gives a lower test WRMSD of 3.589. Most of the test gain is already reached by the refitted no-window control O178 (non-SCF test 3.865 \kcal), which adopts the GAS22-type same-spin polynomial.
\begin{table}[htbp]
\centering\small
\setlength{\tabcolsep}{5pt}
\caption{Performance on the MGCDB84 SyFES split and the 36-set GSCDB137 external-validation subset.}\label{tab:main-comparison}
\begin{tabular}{llrrr}
\toprule
Functional & Mode & MGCDB84 val.$^b$ & MGCDB84 test$^b$ & GSCDB NER$^c$ \\
\midrule
$\omega$B97M-V & SCF & 3.533 & 4.309 & 1.144 \\
GAS22 & non-SCF & 3.462 & 3.599 & 1.212 \\
GAS22 & SCF & 3.470 & 3.589 & 1.190 \\
O159 & non-SCF & 3.332 & 3.818 & 1.141 \\
O159 & SCF & 3.330 & 3.805 & 1.139 \\
O396$^a$ & non-SCF & 3.138 & 3.841 & 1.147 \\
C356$^a$ & non-SCF & 3.363 & 3.897 & --- \\
O178 (no window)$^a$ & non-SCF & 3.481 & 3.865 & 1.189 \\
GAS22 refit (Codex)$^a$ & non-SCF & 3.489 & 3.628 & --- \\
\bottomrule
\end{tabular}
\par\smallskip\noindent\begin{minipage}{0.97\linewidth}\footnotesize\raggedright $^a$Cached search/final-test evaluation; the other MGCDB non-SCF entries are direct full-grid reevaluations. $^b$WRMSD in \kcal; 2,963 validation and 1,152 test reactions. $^c$Dimensionless mean over 36 equally weighted sets (2,734 observations), with published GAS22 parameters except the explicitly labeled Codex refit. Dashes denote unavailable evaluations. GAS22 SCF retains 20 unconverged MGCDB records and two unconverged GSCDB Pd records; energy-convergence qualifications are detailed in Supporting Information Section~S4.\end{minipage}
\end{table}

All O159 MGCDB84 calculations converged under the reported protocol. GAS22 retained 20 unconverged molecular records; excluding the affected reactions changes the split WRMSDs by at most 0.011 \kcal. Category-resolved results show where the gain arises: O159 improves barrier-height test performance relative to \wb, whereas its easy-isomerization error is slightly higher. Easy isomerizations constitute 590 of the 1,152 test reactions, whereas difficult isomerizations contain only eight. Aggregate changes therefore reflect both category performance and the inherited weighting scheme.

The later search-validation gains do not produce the best held-out functional.\cite{kapoor2023leakage} O396, O388, and O376 have test WRMSDs of 3.841, 3.906, and 3.961, respectively, compared with O159's non-SCF 3.818. C356 improves validation by 3.6\% relative to its training-refitted GAS22 reference but worsens test by 7.4\%, mainly through noncovalent-cluster and noncovalent-dimer regressions.

\subsection{External performance on GSCDB137}
On all 36 evaluated GSCDB137 datasets, O159 SCF has mean NER 1.139, compared with 1.144 for the local \wb\ reference. The paired mean difference is approximately $-0.005$ (bootstrap interval spanning zero, signed-rank $p\simeq0.58$), and on the 20 new datasets the mean NER is 1.269 versus 1.290. O159 and the reference therefore perform comparably on this scope, with no significant difference in either direction.\cite{liang2025gold, lakens2020equivalence}

The local reference closely reproduces the published Q-Chem values: the median absolute difference between local and published \wb\ energies is approximately $3.6\times10^{-4}$ \kcal, while four molecules differ by more than 1 \kcal\ because of recorded electronic-state differences. The published and local reference mean NERs over 36 sets are 1.135 and 1.144, respectively. Comparisons against O159 therefore use the local reference consistently. On the 20 new sets, O159's mean NER lies below 15 of the 16 published comparator values, with only the double hybrid $\omega$B97M(2) lower; on the 19 new datasets other than AE11, its mean SCF NER (1.202) is slightly above that of the local \wb\ reference (1.191).

\begin{figure}[htbp]\centering
\includegraphics[width=\linewidth]{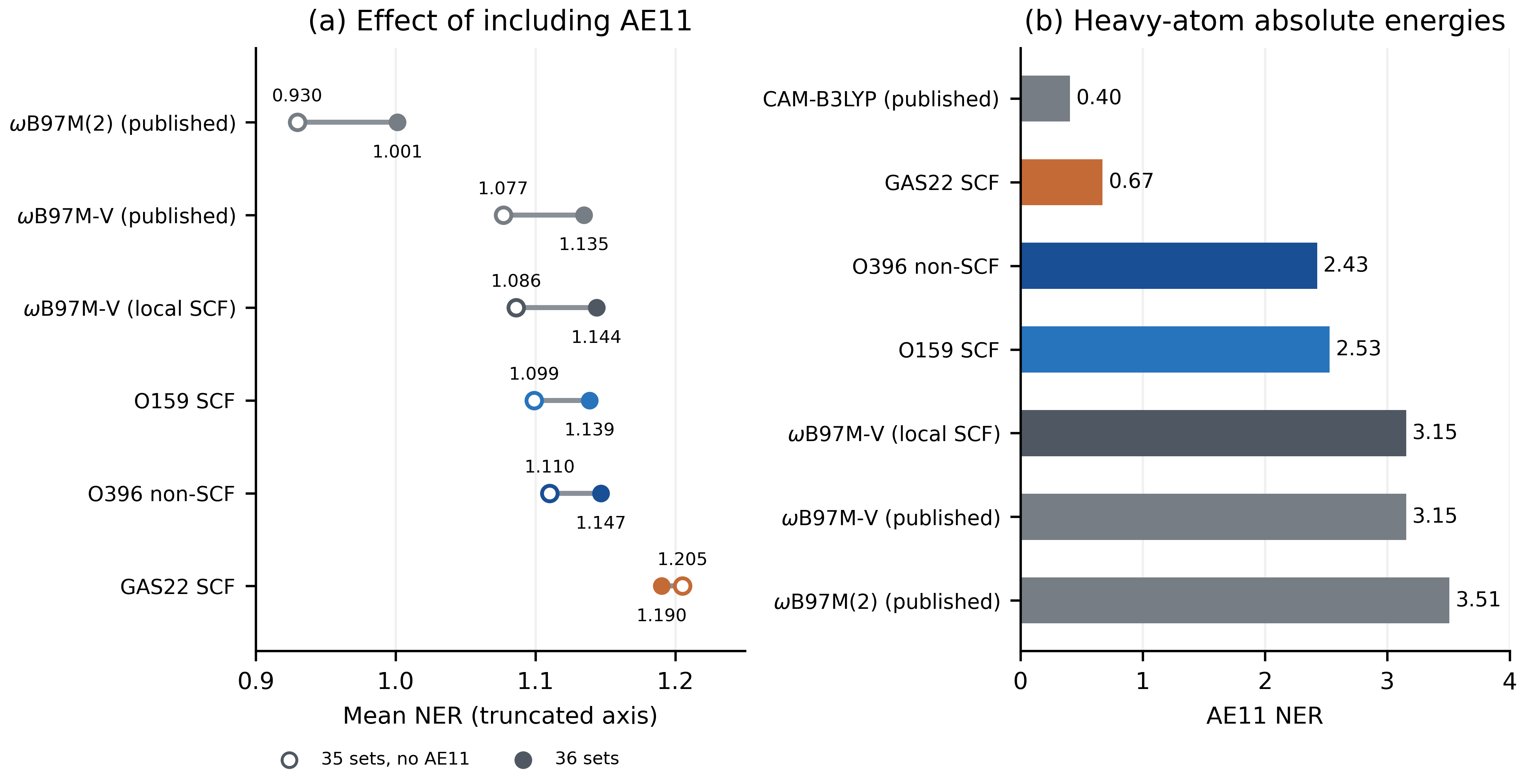}
\caption{External-validation sensitivity to AE11. Left: mean NER for the 35-set scope without AE11 and the completed 36-set scope, with a truncated axis to resolve differences. Right: AE11 alone, an absolute heavy-atom energy dataset. O159 and GAS22 use SCF; O396 is non-SCF. Published reference rows are distinguished from the local \wb\ calculation. AE11 reduces the GAS22--O159 difference substantially, illustrating why the evaluated set membership must accompany aggregate ranks. Scope means are tabulated in Supporting Information Section~S4.}\label{fig:external}
\end{figure}

GAS22 changes relative position between benchmarks. It has the lowest MGCDB84 test error among the evaluated SCF candidates but a higher mean NER than O159 on the external subset. AE11 halves this difference (Figure~\ref{fig:external}): the GAS22-minus-O159 mean NER difference is approximately 0.106 without AE11 and 0.052 with it (36 sets: signed-rank $p\simeq0.020$, bootstrap interval approximately $[-0.085,0.165]$).

Transfer varies with property class. O159 has a lower mean NER than the local reference on the nine noncovalent sets, but higher errors on the evaluated transition-metal, field-response, and vibrational categories. Section~\ref{sec:window-external} separates the contributions of the window and of the refitted coefficients to these differences. O396's lower MGCDB84 validation error does not yield a lower external mean NER than O159 (1.147 versus 1.141, both non-SCF).

\subsection{External test against the refitted control}\label{sec:window-external}
On the 20 new datasets, the prespecified primary scope, 14 improve with the window, but larger losses on a few offset these gains: relative to the refitted no-window control O178, the mean NER changes by $\Delta=-0.041$ (95\% CI $[-0.171,\,0.054]$, $p=0.55$; Table~\ref{tab:window-gscdb}), and by $+0.011$ without AE11. Over all 36 datasets, the window lowers the mean NER from 1.189 to 1.141 (signed-rank $p=0.041$), to the level of the local \wb\ reference (1.144). The most consistent gain is on the 16 datasets with MGCDB84 antecedents ($\Delta=-0.057$, $p=0.018$), which re-test MGCDB84 chemistry with revised references. As for the MGCDB84 controls (Supporting Information Section~S3), each difference measures the window together with the coefficient readjustment it induces.

\begin{table}[htbp]
\centering\small
\setlength{\tabcolsep}{4pt}
\caption{External test of the window against the refitted no-window control O178 on the GSCDB137 subset (non-SCF at \wb\ densities; mean NER, dimensionless). $\Delta$ is O159 minus O178; negative values favor the window.}\label{tab:window-gscdb}
\begin{tabular}{lrrrrrrr}
\toprule
Datasets & $n$ & O178 & O159 & \wb\ & $\Delta$ & 95\% CI & $p$ \\
\midrule
New (primary) & 20 & 1.312 & 1.270 & 1.290 & $-0.041$ & $[-0.171,\,0.054]$ & 0.55 \\
New, without AE11 & 19 & 1.193 & 1.204 & 1.191 & $+0.011$ & $[-0.046,\,0.072]$ & 0.80 \\
With MGCDB84 antecedents & 16 & 1.036 & 0.979 & 0.961 & $-0.057$ & $[-0.109,\,-0.010]$ & 0.018 \\
All & 36 & 1.189 & 1.141 & 1.144 & $-0.048$ & $[-0.121,\,0.011]$ & 0.041 \\
All, without AE11 & 35 & 1.122 & 1.101 & 1.086 & $-0.020$ & $[-0.060,\,0.021]$ & 0.067 \\
\bottomrule
\end{tabular}
\par\smallskip\noindent\begin{minipage}{0.97\linewidth}\footnotesize\raggedright O178 and O159 are non-SCF; \wb\ is the local SCF reference, and O159 SCF means differ from the non-SCF values by at most 0.003. New datasets have no MGCDB84 antecedent. $\Delta$ is computed from unrounded means. Bootstrap intervals resample datasets; $p$ is from the paired signed-rank test. With the window, 26 of 36 datasets improve and 10 do not (14 of 20 new; 12 of 16 with antecedents). The rows are overlapping scopes and are not corrected for multiple comparisons. Per-dataset values and the recorded predictions are in Supporting Information Section~S3.\end{minipage}
\end{table}

The largest external gain is for heavy-atom total energies (AE11, Ca to Rn), whose mean absolute error falls by 29\%, from 415.8 to 295.8 \kcal, the largest NER change of any dataset. Without the window, the refitted frame is less accurate than \wb\ on AE11 (NER 3.555 versus 3.154) and on the 36-set mean; the window brings AE11 below the reference (NER 2.529) and the 36-set mean to its level. This direction was predicted from the training trend for the AE18 atomic total energies (H--Ar; RMSD 7.60 to 5.56 \kcal). For P to Ar, the AE18 improvement comes from the co-adapted coefficients rather than from the window energy itself, and the atomic window energies lie largely outside the low-gradient, near-iso-orbital region ($w_\sigma\geq0.3$) that dominates the reaction-energy attribution (Supporting Information Section~S3). The heavy-atom gain therefore reflects a different aspect of the window from its effect on reaction energies.

The comparison also indicates where O159's external differences from \wb\ originate. The three transition-metal sets on which O159 trails \wb\ (3d4dIPSS, DAPD, and TMD10) are 1.8--6.0\% more accurate with the window than without it, so those losses arise from the refitted base form (the GAS22-type same-spin polynomial and the refitted coefficients) rather than from the window. For V30 and Pol130, O178 (non-SCF) is at least as accurate as SCF \wb, so these losses accompany the window.

Per-dataset results resolve the external effect of the window. Among the new datasets, P34EA electron affinities worsened by 19\%, as predicted from the EA13 and G21EA losses on MGCDB84, whereas G21EA itself improved by 20\% on GSCDB137. The halogen, chalcogen, and pnictogen bonds of HCP32, predicted to improve, worsened by 16\%, and the radical complexes of O24, predicted to worsen, improved by 8\%. Among datasets without a prediction, MB08-165 and the oriented-field set OEEF improve by 14--16\%, whereas vibrational frequencies (V30, $+30$\%) and double ionization energies (DIPCS9, $+22$\%) worsen. The recorded predictions were correct for 11 of 16 antecedent and 3 of 7 new datasets, or for 13 of 16 antecedent datasets when the stated rule is applied to WCPT6 and CT20, whose recorded predictions departed from it (Supporting Information Section~S3); predicting improvement everywhere would have scored 12 of 16 and 5 of 7.

\subsection{Limitations and future directions}
Training uses MGCDB84, and external evaluation covers 36 GSCDB137 datasets that exclude larger systems; extending to the full database would strengthen assessment across chemical regimes.\cite{liang2025gold} Difficult thermochemistry remains a target: O159 improves on GAS22 in the difficult-thermochemistry (TCD) test subset but trails \wb. Broader descriptors, flexible hybrid parameters, and additional exact conditions offer directions for improvement, including tests of fractional-spin behavior.\cite{kaplan2023predictive, ma2022evolving, furness2020accurate, kirkpatrick2021pushing}

 On the new GSCDB137 datasets, only the heavy-atom improvement transferred clearly, and the primary aggregate change was not significant. The descriptor analysis does not explain why the window helps heavy-atom total energies but harms vibrational frequencies and halogen, chalcogen, and pnictogen bonding. Testing the rule in other base functionals, including ones with a Stoll-consistent same-spin factor, would show whether the reduction in magnitude of the same-spin factor at intermediate density is general or specific to this hybrid frame; the single nonhybrid B97M-V control changes within fit noise.

Future work should test periodic SCF evaluation during search\cite{li2021kohn, sim2022improving, kalita2022well} and category-resolved objectives with independent validation partitions, to balance performance across properties and assess generalization.\cite{kapoor2023leakage, bates2024cross} O159 passes all four screening checks and converges across MGCDB84; broader grid and derivative tests would support further applications.\cite{furness2020accurate, bartok2019regularized} Supporting Information Sections~S4 and S2 document external convergence qualifications and an incorrect low-density series in the archived exchange-attenuation implementation, whose correction changes no reported value.

\section{Conclusions}
Frontier LLM agents connected to a shared evaluator produced compact exchange--correlation functionals together with an inspectable record of how they were selected. The choice of backbone mattered: the Opus-5.5 agents reached the SCF gate with about 40-fold fewer tokens than the Codex agents and found the candidate with lower validation and held-out errors, O159, which retains its 11.7\% test improvement over \wb\ under self-consistency and performs comparably to \wb\ on external GSCDB137 data.

Beyond a single functional, the search archive supplied the refitted controls that turn the discovered term into a design rule: in this hybrid frame, a same-spin density window should scale the entire same-spin factor at intermediate density, without acting on opposite-spin correlation or through a kinetic-energy gate. The agents' advantage over SyFES evolution at a matched number of fits came from an explicit density scale that scale-invariant B97-type series lack, which points to density-dependent corrections of these series as a productive direction for functional design. More broadly, coupling language-model agents to physics-based screening, self-consistent evaluation, and external benchmarks makes automated search a source of testable design hypotheses as well as candidate functionals, with every proposal, control, and failure recorded for others to audit and extend.

\section*{Technology Use Disclosure}
ChatGPT and Claude were utilized for grammatical refinement and typographical corrections. The authors have carefully reviewed and verified all sections of the manuscript for accuracy and integrity.

\section{Data availability}

The source code is publicly available on GitHub at
\url{https://github.com/BaratiLab/DFT_functional_agentic_searcher}. Supporting Information Section~S8 describes the contents of the reproducibility archive and its verification.

\section*{Author Contributions} A.C. performed software development and testing and did benchmarking. A.C. carried out the result analysis. Writing of the manuscript was done by A.C. ; A.B.F. conceived and approved the manuscript.

\section*{Conflicts of interest}
\label{sec:conflicts}
The authors declare no conflicts of interest.

\begin{acknowledgement}

The authors would like to acknowledge the Mechanical and AI Lab at Carnegie Mellon University for computational resources.

\end{acknowledgement}

\begin{suppinfo}
The Supporting Information is available free of charge: computational methods (S1), functional equations and parameters (S2), density-window analysis (S3), performance comparisons (S4), the SyFES baseline (S5), token accounting (S6), agent workflow and representative exchanges (S7), and data and reproducibility (S8).
\end{suppinfo}

\begin{tocentry}
\centering
\includegraphics[width=3.25in,height=1.75in,keepaspectratio]{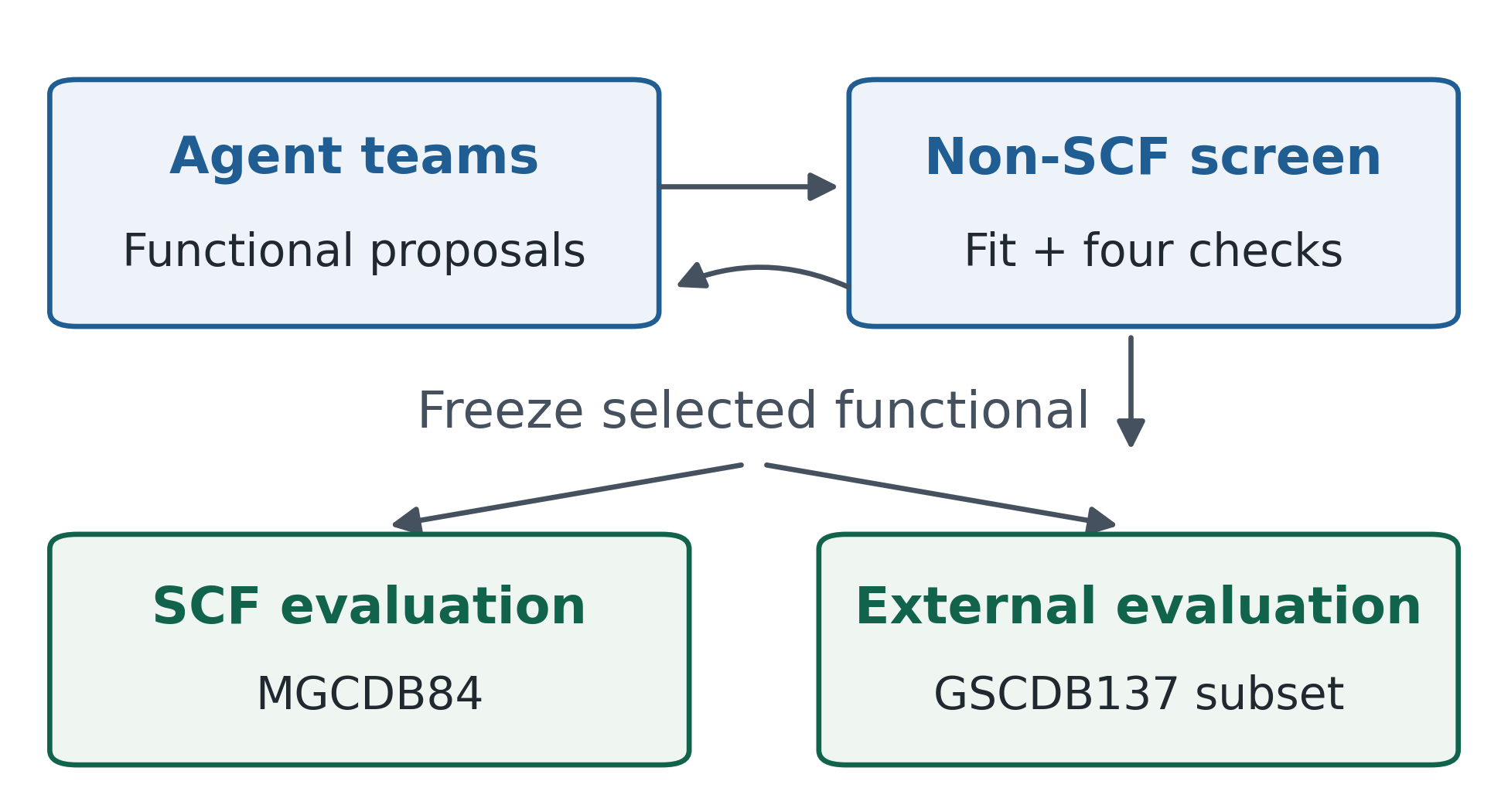}
\par\vfill
\end{tocentry}

\clearpage
\setcounter{section}{0}
\setcounter{subsection}{0}
\setcounter{table}{0}
\setcounter{figure}{0}
\setcounter{equation}{0}
\setcounter{page}{1}
\renewcommand{\thepage}{S\arabic{page}}
\renewcommand{\thesection}{S\arabic{section}}
\renewcommand{\thetable}{S\arabic{table}}
\renewcommand{\thefigure}{S\arabic{figure}}
\renewcommand{\theequation}{S\arabic{equation}}
\renewcommand{\theHsection}{S\arabic{section}}
\renewcommand{\theHtable}{S\arabic{table}}
\renewcommand{\theHfigure}{S\arabic{figure}}
\renewcommand{\theHequation}{S\arabic{equation}}
\raggedbottom
\begin{center}
{\Large\bfseries Supporting Information}\par\vspace{0.6em}
{\Large\bfseries\papertitle}\par\vspace{1em}
Achuth Chandrasekhar$^{a}$ and Amir Barati Farimani$^{a,b,c,d,e,*}$\par\vspace{0.7em}
\small
$^{a}$Mechanical Engineering; $^{b}$Biomedical Engineering;\par
$^{c}$Chemical Engineering; $^{d}$Materials Science and Engineering;\par
$^{e}$Machine Learning Department, Carnegie Mellon University,\par
Pittsburgh, PA 15213, USA\par
$^{*}$Correspondence: \href{mailto:barati@cmu.edu}{barati@cmu.edu}
\end{center}
\vspace{1em}
Essential methods, functional definitions, performance comparisons, and search records are presented here. O159 and O396 denote Opus database records 159 and 396; C356 denotes Codex record 356. Complete iteration ledgers, parameter sets, dataset membership, and numerical records are supplied in the accompanying reproducibility archive.
\clearpage
\section{Computational methods}\label{sec:si-methods}
\subsection{Datasets and objectives}
The searches used MGCDB84 reaction definitions and geometries from ACCDB, with the released SyFES partition and weighting procedure.\cite{mardirossian2017thirty,ma2022evolving,morgante2019accdb} Omitting XB51\_15 leaves 4,985 reactions and 5,930 unique molecular calculations. The training, validation, and test partitions contain 870, 2,963, and 1,152 reactions, respectively. Parameters were fitted only on training data; validation guided functional-form selection. Subsets were assigned as groups, so the partitions differ in chemical composition. In particular, the test partition contains 590 easy-isomerization reactions but only eight difficult-isomerization reactions.

For partition $S$, weighted root-mean-square deviation (WRMSD) is
\begin{equation}
J_S=\left[\frac{1}{N_S}\sum_{i\in S}w_i
(\Delta E_i-\Delta E_i^{\rm ref})^2\right]^{1/2},\qquad
\Delta E_i=\sum_m\nu_{im}E_m ,\label{eq:wrmsd}
\end{equation}
where $\nu_{im}$ is the signed stoichiometric coefficient and energies are in \kcal\ ($1\Eh=627.509474$ \kcal). The denominator is the reaction count, not the sum of weights. Category weights are 0.1 for difficult thermochemistry; 1 for easy thermochemistry and AE18; 10 for difficult noncovalent dimers, difficult isomerizations, and barrier heights; 100 for easy noncovalent dimers/clusters and easy isomerizations; and 10,000 for RG10. Subset normalization follows SyFES, pooling TCE with AE18 and NCED with NCEC. Positive-reference RG10 entries receive an additional factor $10^{-4}$, and validation weights are doubled. Exact reaction-level weights and membership are archived.

The combined diagnostic $J_{\rm VT\setminus RG10}$ pools validation and test, excluding RG10 and retaining their existing weights (3,546 reactions). It includes selection data and is not an independent test metric. A post-search reporting variant, \texttt{agentic2026}, moves all eight C20C24 test reactions to validation, giving counts 870/2,971/1,144. This approximates the SAFS26 partition but does not reproduce its unspecified two-molecule transfer exactly.\cite{duston2026agentic} The discovery runs used the original SyFES partition.

\subsection{Reference densities, fitting, and screening}
Reference orbitals used \wb/def2-QZVPPD, restricted or unrestricted Kohn--Sham calculations as appropriate, GPU4PySCF, and def2-universal-jkfit density fitting.\cite{head2016v,sun2020recent,li2025introducing,rappoport2010property} The semilocal grid was PySCF level 3; VV10 used SG-1 pruning for systems with nuclear charges at most 18 and the default nonlocal grid otherwise. Reference SCF tolerance was $10^{-10}\Eh$. All 5,930 systems converged, requiring 84.6 GPU-h on three NVIDIA RTX 6000 Ada GPUs. Final reference energies used exact Coulomb/exchange integrals at the density-fitted orbital solution.

For candidates in the same hybrid frame, fixed-density screening replaces the semilocal energy:
\begin{equation}
E^F_{\rm nonSCF}=E^0+E_{\rm sl}^{F}[\rho^0]
-E_{\rm sl}^{0,\mathrm{torch64}}[\rho^0].\label{eq:nonscf}
\end{equation}
The exact-exchange difference is additionally removed for the nonhybrid B97M-V seed. Cached features omit points with total density below $10^{-11}$ a.u. and are stored in float32. The reference semilocal baseline is computed before that conversion; promoting cached features to float64 does not recover their original precision. Accordingly, cached search and full-grid reevaluation values are kept distinct: the \wb\ validation baseline is 3.533590004 in search metadata and 3.533226799 from the full-grid energies.

Every retained empirical coefficient and introduced empirical shape parameter was jointly fitted on training $J^2$ using PyTorch L-BFGS with strong-Wolfe line search, history 30, gradient tolerance $10^{-10}$, and objective-change tolerance $10^{-12}$. Restarts were selected by training loss. The search prompt limited requests to three restarts and 500 optimizer iterations; the server imposed a 560 s tool timeout and usual 410 s fitting budget. Training used compiled float32 kernels with an eager fallback; final metrics used float64. Budget exhaustion retained the best training iterate and did not certify optimization convergence. The recorded environment includes Python 3.12.14, PySCF 2.14.0, GPU4PySCF 1.8.1, PyTorch 2.14.1, CuPy 14.2.0, and MCP 1.30.0.

Four screening checks were applied:
\begin{itemize}
\item \textbf{Spin interchange:} maximum $|e_{xc}(\alpha,\beta)-e_{xc}(\beta,\alpha)|/\rho<10^{-5}\Eh$ on triplet O$_2$ points with $\rho>10^{-6}$ a.u.
\item \textbf{Uniform-gas exchange:} absolute $|g_x-0.85|<0.005$ at unpolarized $r_s=0.5,1,2,5$, after removing LDA exchange and range attenuation; the nonhybrid target is one.
\item \textbf{Exchange-factor scaling:} energy-weighted integrated change below $10^{-4}\Eh$ on water under coordinate scales 0.5, 2, and 5, with range attenuation divided out.
\item \textbf{Grid sensitivity:} maximum semilocal-energy difference below 0.015 \kcal\ across AE18 atoms on fixed densities integrated with (99,590) and (250,974) radial/angular grids.
\end{itemize}
Nonfinite outputs were rejected. The score was $R=(J_{\rm target}/J_{\rm val})0.9^{n_v}$, with stored $J_{\rm target}=3.032558586$ \kcal\ and violation count $n_v$. Routine SCF eligibility required zero violations and validation below 97\% of the cached published \wb\ baseline. Passing these finite tests is not equivalent to satisfying all exact functional constraints or demonstrating SCF stability.

\subsection{SCF and external evaluation}
Frozen-parameter SCF calculations use PyTorch automatic differentiation to supply derivatives with respect to density, density gradients, and kinetic-energy density to GPU4PySCF. MGCDB84 candidate SCFs start from the reference orbitals, use a $10^{-9}\Eh$ energy tolerance, and receive the exact-integral final-energy correction. Only O159 and published-parameter GAS22 have complete candidate MGCDB84 SCF campaigns. The implementation supplies first derivatives, not analytic Hessians or response kernels.

External evaluation followed the molecule-specific GSCDB137 Q-Chem inputs, including basis sets, grids, effective core potentials, electric fields, and electronic-state checks.\cite{liang2025gold} The cost cutoff selected 36 datasets whose largest molecule has at most 500 basis functions: 2,734 entries and 3,487 molecular/field calculations. Sixteen datasets have MGCDB84 antecedents; 20 do not. GSCDB137 was used for reporting only, with no fitting or functional selection. A complete SCF evaluation over the entire GSCDB137 dataset was not performed because of computational resource constraints.

Density fitting was restricted to def2 bases; other bases used exact Coulomb/exchange throughout. Documented high-angular-momentum issues required CPU core Hamiltonians and separate full-/long-range exchange builds. AE11 Rn was completed on the CPU and is included in the 36-set results. GSCDB mean normalized error ratio (NER) is
\begin{equation}
\overline{\NER}_D(F)=\frac1{|D|}\sum_{d\in D}\frac{M_d(F)}{s_d},
\end{equation}
where $M_d$ is the benchmark-defined error measure and $s_d$ its fixed published normalization. Most sets use mean absolute error; the property datasets use their specified relative, weighted, or frequency metrics. NER is dimensionless and weights datasets equally, unlike reaction-weighted MGCDB84 WRMSD. Statistical comparisons use paired datasets, signed-rank tests, and 10,000-replicate bootstrap intervals (20,000 for the window control of Section~\ref{sec:si-window-control}); reported intervals and their reproducible reanalysis are retained in the archive.

\section{Principal functional expressions}\label{sec:si-functionals}
\subsection{Shared definitions and reference forms}
Atomic units are used, with energy densities per unit volume. For spin $\sigma\in\{\alpha,\beta\}$, define
\begin{align}
X_\sigma&=|\nabla\rho_\sigma|^2/\rho_\sigma^{8/3},&
u_\gamma(X)&=\frac{\gamma X}{1+\gamma X},\\
\tau_\sigma&=\tfrac12\sum_i n_{i\sigma}|\nabla\phi_{i\sigma}|^2,&
\tau_{U\sigma}&=\tfrac3{10}(6\pi^2)^{2/3}\rho_\sigma^{5/3},\\
\tau_{W\sigma}&=|\nabla\rho_\sigma|^2/(8\rho_\sigma),&
t_\sigma&=\tau_{U\sigma}/\tau_\sigma,\quad w_\sigma=\frac{t_\sigma-1}{t_\sigma+1},\\
r_{s\sigma}&=(3/4\pi\rho_\sigma)^{1/3},&r_s&=(3/4\pi\rho)^{1/3}.
\end{align}
Here $\phi_{i\sigma}$ and $n_{i\sigma}$ are orbitals and occupations, $\rho=\rho_\alpha+\rho_\beta$, and $\zeta=(\rho_\alpha-\rho_\beta)/\rho$. The default opposite-spin variables are $\bar X=(X_\alpha+X_\beta)/2$, $\bar t=(t_\alpha+t_\beta)/2$, and $\bar w=(\bar t-1)/(\bar t+1)$; thus $\bar w$ is not generally the mean of the two $w_\sigma$. Inactive channels are masked. Radii inside empirical factors denote their numerical values in bohr.

All principal candidates use the fixed range-separated hybrid frame
\begin{align}
E_{xc}={}&\int\left[\sum_\sigma e_{x\sigma}^{\rm LDA}F(a_\sigma)g_{x\sigma}
+\sum_\sigma e_{c\sigma\sigma}^{\rm PW}g_{ss,\sigma}+e_{c\alpha\beta}^{\rm PW}g_{os}\right]d\mathbf r\nonumber\\
&+0.15E_{x,\rm SR}^{\rm HF}+E_{x,\rm LR}^{\rm HF}+E_c^{\rm VV10},\label{eq:frame}
\end{align}
with $\omega=0.3\,a_0^{-1}$ and VV10 parameters $b=6$, $C=0.01$.\cite{head2016v,vydrov2010nonlocal} Short-/long-range exact exchange use the erfc/erf kernels, respectively. Local exchange is $e_{x\sigma}^{\rm LDA}=-(3/8)(3/\pi)^{1/3}(2\rho_\sigma)^{4/3}$; $a_\sigma=\omega/(6\pi^2\rho_\sigma)^{1/3}$. Correlation uses the Stoll partition of PW92 with the implementation's \texttt{PW\_MOD} constants:\cite{perdew1992accurate,stoll1978calculation}
\begin{equation}
e_{c\sigma\sigma}^{\rm PW}=\rho_\sigma\epsilon_c(r_{s\sigma},1),\quad
e_{c\alpha\beta}^{\rm PW}=\rho\epsilon_c(r_s,\zeta)-\sum_\sigma e_{c\sigma\sigma}^{\rm PW}.
\end{equation}
These established local and nonlocal ingredients are implemented in the archived source; only the enhancement factors are evolved. The semilocal exchange constant 0.85 already appears in $g_x$ and must not be applied again outside it.

For reference, the \wb\ factors have the polynomial bases\cite{head2016v}
\begin{align}
g_x&=0.85+a_0w_\sigma+a_1u_{0.004}(X_\sigma),\\
g_{ss,\sigma}&=b_0+b_1w_\sigma+b_2w_\sigma^2+b_3w_\sigma^4u_s^3+b_4u_s^4,\\
g_{os}&=c_0+c_1\bar w+c_2\bar w^2+c_3\bar w^6+c_4\bar w^2u_o+c_5\bar w^6u_o,
\end{align}
where $u_s=u_{0.2}(X_\sigma)$ and $u_o=u_{0.006}(\bar X)$. The training refit varies 13 coefficients. GAS22 instead uses\cite{ma2022evolving}
\begin{align}
g_x&=a_0+a_1u_{\gamma_x}(X_\sigma)+a_2w_\sigma,\\
g_{ss,\sigma}&=u_s+b_0w_\sigma+b_1w_\sigma^2+b_2u_s^6+b_3u_s^6w_\sigma^4,\\
g_{os}&=c_0+c_1\bar w^2+c_2\bar w^2Q+c_3\bar w^6+c_4\bar w^6Q,
\end{align}
with $\gamma_x=0.003840616724010807$, $\gamma_s=0.46914023462026644$, $u_s=u_{\gamma_s}(X_\sigma)$, and $Q=(\bar X+10^{-30})^{1/3}$. Its 12 published coefficients are distinct from the training-refitted seeds. Published GAS22 has $g_x=0.862139736374172$ at the uniform gas and fails the specified 0.85 check. Reference coefficients and all evaluated refits accompany the executable source.

\subsection{O159: same-spin density window}
Let $h_p(r)=r/(1+r^p)$ and
\begin{equation}
P_s(w,u)=u+b_0w+b_1w^2+b_2u^6+b_3u^6w^4.
\end{equation}
The complete O159 factors are
\begin{align}
g_{x\sigma}&=0.85+a_0w_\sigma+a_1u_{0.004}(X_\sigma),\\
g_{ss,\sigma}&=P_s(w_\sigma,u_{\gamma_s}(X_\sigma))
\left[1+(\beta_0+\beta_2w_\sigma^2)h_4(r_{s\sigma})\right],\label{eq:window}\\
g_{os}&=1+c_0\bar w+c_1\bar w^2+c_2\bar w^6+c_3\bar w^2u_o+c_4\bar w^6u_o,
\end{align}
where $u_o=u_{0.006}(\bar X)$. All 13 retained coefficients were jointly refitted; retaining a reference polynomial form does not mean freezing its published coefficients. Table~\ref{tab:o159-coefficients} gives the evaluated values; full precision is stored in \texttt{functional\_parameters.json}.

\begin{table}[htbp]\centering\small
\caption{O159 fitted coefficients, rounded to eight decimal places.}\label{tab:o159-coefficients}
\begin{tabular}{lrlr}\toprule
Parameter & Value & Parameter & Value\\\midrule
$a_0$ & 0.24549733 & $c_0$ & 1.14502582\\
$a_1$ & 1.00304439 & $c_1$ & 3.75266635\\
$b_0$ & -4.69984961 & $c_2$ & -3.35815548\\
$b_1$ & -4.88244588 & $c_3$ & -10.59006317\\
$b_2$ & -1.96642642 & $c_4$ & 10.37863864\\
$b_3$ & 7.80561285 & $\beta_0$ & -0.25463509\\
 & & $\beta_2$ & -0.26628913\\\bottomrule
\end{tabular}
\end{table}

\begin{figure}[htbp]\centering
\includegraphics[width=0.90\linewidth]{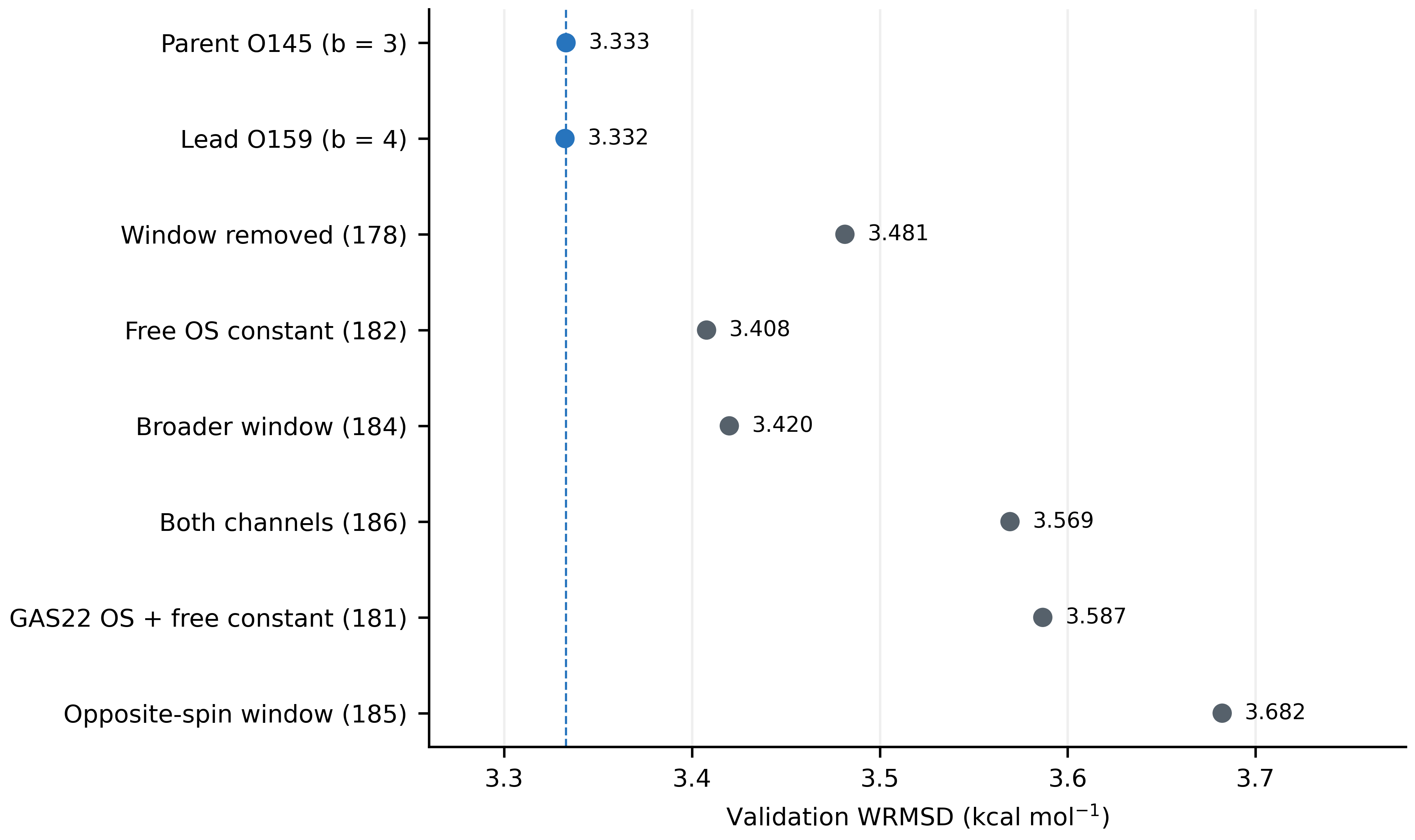}
\caption{Refitted ablations of the O145 density-window architecture. Each point uses a separate joint training fit; the dashed line denotes O145 validation error. Removing the window raises validation WRMSD by approximately 0.148 \kcal, while changing the cubic cutoff to the quartic O159 form gives nearly the same validation error. These comparisons do not isolate a term's contribution at fixed coefficients. The analytic window shape is shown in the main manuscript rather than repeated here.}\label{fig:window-analysis}
\end{figure}

The window approaches unity at both density extremes and changes the same-spin factor at intermediate density, as illustrated in the main manuscript's density-window panel. Its parent O145 uses $h_3$ with independently fitted parameters. Figure~\ref{fig:window-analysis} provides the complementary refitted ablation analysis: changing to $h_4$ improved validation by only 0.000562 \kcal, whereas removing the window worsened validation by about 0.148 \kcal. The exponent is therefore not determined by the data, whereas the window's channel and its scope (the whole same-spin factor) are; profiles that remain large near $r_{s\sigma}\approx2$ bohr are indicatively disfavored (Section~\ref{sec:si-window}).

\subsection{O396 and C356: later validation-selected candidates}
For O396 define
\begin{equation}
A_\sigma=\max[(\tau_\sigma-\tau_{W\sigma})/\tau_{U\sigma},0],\quad
L_\sigma=\frac{A_\sigma^2}{1+A_\sigma^2},\quad \bar L=\tfrac12(L_\alpha+L_\beta).
\end{equation}
With $u_s=u_{\gamma_s}(X_\sigma)$, $\bar u_s=u_{\gamma_s}(\bar X)$, and $u_o=u_{0.006}(\bar X)$, its ten-parameter form is
\begin{align}
g_x={}&0.85+a_0w_\sigma+a_1u_{0.004}(X_\sigma),\\
g_{ss,\sigma}={}&[u_s+b_0w_\sigma(1+w_\sigma)+b_1u_s^6+b_2u_s^6w_\sigma^4]
[1+\beta(1+w_\sigma^2)h_4(r_{s\sigma})]\nonumber\\
&+100s_s h_4(r_{s\sigma})L_\sigma^3(1-u_s)^8,\\
g_{os}={}&1+\bar w+\bar w^2(1-\bar w^4)(c_0+c_1u_o)
+100s_o h_4(2^{1/3}r_s)\bar L^3(1-\bar u_s)^8.
\end{align}
The stored shell coefficients are $s_s=2.7791162414840516$ and $s_o=-1.9737450397311325$; the factor 100 is explicit. O388 omits the last opposite-spin shell term and has nine independently fitted parameters. The shell-region hypothesis was informed by a training-plus-validation residual scan, so its validation gain is not independent evidence for the proposed mechanism.

For C356, define the distinct density-weighted kinetic descriptor and smoothed root response
\begin{align}
Z_\sigma&=\operatorname{clip}_{[0,1]}\left(\frac{\tau_{W\sigma}}{\tau_\sigma+10^{-30}}\right),&
W&=\tfrac{1+\zeta}{2}w_\alpha+\tfrac{1-\zeta}{2}w_\beta,\\
s&=\operatorname{softplus}(h)+10^{-12},&R&=(\bar X+s)^{1/3}-s^{1/3},\\
V&=R/(1+0.006R),&\lambda&=\operatorname{softplus}(q),
\end{align}
where $\operatorname{softplus}(y)=\ln(1+e^y)$. Its 14 fitted parameters enter
\begin{align}
g_x={}&0.85+a_0u_{\gamma_x}(X_\sigma)+a_1w_\sigma,\\
g_{ss,\sigma}={}&b_0u_s+b_1w_\sigma(1+w_\sigma)+b_2u_s^6+b_3u_s^6w_\sigma^4
+b_4Z_\sigma(1-Z_\sigma),\\
g_{os}={}&c_0+c_1W^2(1-W^4)+c_2W^2V+c_3W^6V
+K\frac{W(1+W)}{\sqrt{1+(\lambda r_s)^2}}.
\end{align}
Only the pure $W^2/W^6$ pair is tied; the root couplings remain independent. The orbital-window term vanishes at $Z=0,1$, but does not force the entire same-spin correlation to vanish for one electron. The density screen uses total-density $r_s$, unlike the O159 spin-channel radius. Neither O396 nor C356 has an SCF result; C356 was not evaluated on GSCDB137. Their evaluated coefficient sets and precursor/shortlist variants are supplied in the archive rather than repeated as long parameter tables here.

\subsection{Numerical guards and the attenuation branch}\label{sec:si-attenuation}
The standard implementation activates a spin channel at $\rho_\sigma>10^{-10}$ a.u., floors squared gradients at $10^{-20}$, and imposes $\tau_\sigma\ge\max(10^{-14},\tau_{W\sigma})$. Inactive channels are masked from energies and default opposite-spin averages. GSCDB convergence repairs lowered the density threshold to $10^{-14}$ a.u. for the repaired records; the archive preserves that exception.

The analytic short-range exchange attenuation is
\begin{equation}
F(a)=1-\frac23a\left[2\sqrt\pi\,\operatorname{erf}(1/a)-3a+a^3
+(2a-a^3)e^{-1/a^2}\right],
\end{equation}
with $a=\omega/k_{F\sigma}$ and $k_{F\sigma}=(6\pi^2\rho_\sigma)^{1/3}$. To avoid cancellation, the archived PyTorch implementation switched at $a\ge5$ to the series $1/(36a^2)-1/(960a^4)+1/(26880a^6)$. That series is the large-$a$ expansion for the alternative convention $a'=\omega/(2k_{F\sigma})$; in the present variable, it equals $F(2a)$ rather than $F(a)$. The correct expansion is $1/(9a^2)-1/(60a^4)+1/(420a^6)$, which matches the closed form to $8\times10^{-10}$ (relative $2\times10^{-7}$) at $a=5$. The archived branch was therefore four times too small and discontinuous at $a=5$ (0.0011094 versus 0.0044179). At $\omega=0.3\,a_0^{-1}$, it affects only spin densities $10^{-10}<\rho_\sigma<3.65\times10^{-6}$ a.u. The upstream SyFES code evaluates the closed form throughout, so the published GAS22 parameters are unaffected. All archived O159, O396, C356, and GAS22 energies, fits, and SCF calculations used the branch, whereas the \wb\ reference energies were computed with libxc and are unaffected. The same branch also entered the torch \wb\ energy that the non-SCF expression subtracts.

We quantified the effect with a corrected-kernel reevaluation that used the archived parameters, reference densities, and integration grids, without refitting or new SCF calculations (Table~\ref{tab:si-attenuation}). Non-SCF energies were recomputed as $E+\Delta_{\mathrm{cand}}-\Delta_{\mathrm{ref}}$, where $\Delta$ is the corrected-minus-archived semilocal energy of the candidate or of torch \wb\ at the same density. Cancellation between the two terms makes the non-SCF shifts very small. For SCF energies, the candidate shift $\Delta_{\mathrm{cand}}$ was added at the \wb\ density. Two approximations are involved. Because the SCF energy is variational, the density response to the correction contributes only at second order. Evaluating the shift at the \wb\ density rather than at the candidate's own density introduces an error proportional to the product of two small quantities, the shift itself and the density difference. Both are negligible at the reported precision. On MGCDB84, about 10\% of the stored spin-channel grid points lie in the affected range. These are tail points: the largest per-molecule candidate shift is $2.6\times10^{-4}$ \kcal\ (C$_{24}$ ring, C20C24), and the largest non-SCF reaction-energy change is $5.7\times10^{-5}$ \kcal\ (GAS22). On the 3,487 GSCDB137 calculations, the largest per-molecule shift is $4.6\times10^{-4}$ \kcal. The most sensitive dataset is V30, whose frequencies are second finite differences of energies; its metric changes by at most 0.12\% (GAS22 SCF, 32.021 to 32.059). Every WRMSD, mean NER, ranking, paired difference, and signed-rank $p$ value reported in this work is unchanged at its reported precision. The O159--\wb\ mean NER difference over 36 sets changes from $-0.00492$ to $-0.00489$ ($p=0.58$ in both cases), and the GAS22--O159 difference changes from 0.05158 to 0.05160 ($p=0.020$).

\begin{table}[htbp]
\centering\small
\setlength{\tabcolsep}{4pt}
\caption{Effect of correcting the large-$a$ attenuation branch, using archived parameters and densities. MGCDB84 WRMSD changes are in \kcal; GSCDB137 values are mean NER changes over 36 sets and over the 20 sets without MGCDB84 antecedents. SCF entries apply the first-order shift described in the text.}\label{tab:si-attenuation}
\begin{tabular}{llrrrrr}
\toprule
Functional & Mode & $\Delta$ train & $\Delta$ val. & $\Delta$ test & $\Delta$NER$_{36}$ & $\Delta$NER$_{20}$ \\
\midrule
O159 & non-SCF & $-2\times10^{-8}$ & $-4\times10^{-7}$ & $+7\times10^{-8}$ & $+3.1\times10^{-5}$ & $+6.2\times10^{-5}$ \\
O159 & SCF & $-4.7\times10^{-6}$ & $-6.4\times10^{-5}$ & $+5.6\times10^{-6}$ & $+3.2\times10^{-5}$ & $+6.4\times10^{-5}$ \\
GAS22 & non-SCF & $-3\times10^{-7}$ & $-9\times10^{-7}$ & $-3.8\times10^{-6}$ & $+5.0\times10^{-5}$ & $+8.0\times10^{-5}$ \\
GAS22 & SCF & $+8.6\times10^{-6}$ & $+8.7\times10^{-6}$ & $+4.6\times10^{-5}$ & $+5.1\times10^{-5}$ & $+8.1\times10^{-5}$ \\
O396 & non-SCF & $-2\times10^{-8}$ & $-3\times10^{-7}$ & $+9\times10^{-8}$ & $+4.0\times10^{-5}$ & $+7.4\times10^{-5}$ \\
C356 & non-SCF & $-6\times10^{-8}$ & $-3.7\times10^{-6}$ & $+6.4\times10^{-6}$ & --- & --- \\
\wb\ (libxc) & SCF & 0 & 0 & 0 & 0 & 0 \\
\bottomrule
\end{tabular}
\end{table}

The fitted parameters are equally insensitive to the correction. Starting from the archived O159 coefficients, training-set reoptimization with the archived and with the corrected kernel converged to coefficient sets that differ by at most $4.4\times10^{-6}$. The corresponding validation WRMSDs are 3.332431 and 3.332403 \kcal, and the test WRMSDs are 3.817962 and 3.817956 \kcal. Repeated fits from the original initialization are not bitwise reproducible, because they use compiled float32 kernels. Under both kernels, they reached the same two basins (validation 3.3326--3.3382 \kcal), with a within-kernel coefficient spread of 0.11. This optimizer variability exceeds the kernel effect by more than four orders of magnitude. The O159 coefficients of Table~\ref{tab:o159-coefficients} can therefore be used with the correct attenuation without changing any reported result. All four screening checks pass under both kernels. Automatic differentiation of the archived piecewise energy omits the surface term from the jump at $a=5$, so the archived potential was not strictly stationary there. That term acts only at $\rho_\sigma=3.65\times10^{-6}$ a.u., where $F(a)\leq0.0044$ and the exchange contribution is attenuated below 0.5\%; its effect on SCF energies enters through the same tail densities whose shifts are bounded above. The potential discontinuity at $a=5$ lies at much higher density than the far minority-spin tail ($\rho_\sigma\approx10^{-10}$ a.u.) that was responsible for the GSCDB convergence repairs. O159 and the other candidates are defined with the closed-form $F(a)$. Because the closed form cancels catastrophically for large $a$, it should be evaluated with the corrected series for $a\geq5$; the corrected series is available in the released code, and the archived branch is retained only to reproduce the reported numbers. For the GSCDB137 records repaired with the $10^{-14}$ a.u. density threshold, the affected range extends below $10^{-10}$ a.u.; these records are included in the reevaluation.
\FloatBarrier

\section{Density-window analysis}\label{sec:si-window}
This section derives an empirical design rule for the O159 density window from the refitted controls recorded during the search, localizes its effect on reaction energies, and tests it on external data. MGCDB84 quantities are computed from archived parameters on the cached \wb/def2-QZVPPD reference densities (float32 features, float64 arithmetic); GSCDB137 quantities use the stored \wb\ orbitals on the benchmark's production grids. No functional was refitted and no SCF calculation was performed for this analysis; the $\beta_2$-basin result of Section~\ref{sec:si-design-rule} uses the O159 refits of the attenuation check (Section~\ref{sec:si-attenuation}). Because the attenuation correction of Section~\ref{sec:si-attenuation} changes no reported value, the archived kernel was used.

\subsection{Refitted controls}
Table~\ref{tab:si-window-controls} lists the archived controls. Each is a separate joint training fit proposed and recorded by the agents during the search, usually initialized from its parent. Records O92 and O125 head earlier window lineages on the GAS22 and \wb\ base forms, respectively; O145 combines published-form \wb\ exchange and opposite-spin factors (with the opposite-spin constant fixed at one) with the GAS22 same-spin polynomial.

Because all coefficients are refitted together, the window's contribution cannot be read from its fixed-coefficient energy. Removing the window from O159 while retaining the other coefficients raises validation WRMSD from 3.332 to 5.595 \kcal\ and test WRMSD from 3.818 to 7.188 \kcal, whereas a refit without the window (O178) recovers to 3.481 and 3.865 \kcal. Refitted controls are therefore the relevant comparison (Tables~\ref{tab:si-window-controls} and \ref{tab:window-controls}). The fits depend on their starting point: the O145 form reaches validation WRMSDs of 3.333, 3.386, and 3.425 \kcal\ from three initializations (O145, O188, O187), the last starting from the no-window optimum, and the O152 form spans 3.344--3.392 \kcal. We therefore regard changes larger than this 0.09 \kcal\ spread as robust, changes of 0.05--0.09 \kcal\ as indicative, and smaller changes as within noise. Because the controls were warm-started from their parents, we apply this single-fit spread rather than the $\sqrt2$-larger spread expected between independently initialized fits (0.13 \kcal), which would reclassify only O189 as indicative.

\begin{table}[htbp]\centering\small
\setlength{\tabcolsep}{3pt}
\caption{Refitted controls of the same-spin window (non-SCF, cached search metrics, \kcal). $\Delta$ is the change in validation WRMSD relative to the parent.}\label{tab:si-window-controls}
\begin{tabular}{p{0.15\linewidth}p{0.40\linewidth}rrrr}\toprule
Record (parent) & Change relative to parent & Params & Train & Val. & $\Delta$\\\midrule
O145 (O143) & reference architecture, $h_3$ window & 13 & 3.226 & 3.333 & ---\\
O159 (O145) & $h_3\to h_4$ & 13 & 3.252 & 3.332 & $-0.001$\\
O178 (O145) & window removed & 11 & 3.386 & 3.481 & $+0.148$\\
O185 (O178) & analogous window ($r_s$, $\bar w$) on opposite-spin channel only & 13 & 3.172 & 3.682 & $+0.201$\\
O186 (O178) & one shared window on both channels & 13 & 3.181 & 3.569 & $+0.088$\\
O114 (O92) & total-density $r_s$ instead of $r_{s\sigma}$ & 16 & 3.199 & 3.453 & $+0.074$\\
O136 (O125) & window on the uniform-gas same-spin term only & 14 & 3.323 & 3.667 & $+0.330$\\
O184 (O145) & broad profile $r/(1+r^2)$ & 13 & 3.226 & 3.420 & $+0.087$\\
O173 (O145) & exponential cutoff $re^{-r}$ & 13 & 3.210 & 3.391 & $+0.058$\\
O150 (O145) & profile $2r^2/(1+r^4)$ & 13 & 3.222 & 3.367 & $+0.034$\\
O137 (O125) & $h_3(r_{s\sigma}/0.75)$, peak moved from 0.79 to 0.60 bohr & 14 & 3.246 & 3.352 & $+0.015$\\
O138 (O125) & $h_3(r_{s\sigma}/1.3)$, peak moved from 0.79 to 1.03 bohr & 14 & 3.150 & 3.420 & $+0.083$\\
O96 (O92) & trainable peak scale $r_0$ & 17 & 3.178 & 3.379 & $-0.000$\\
O152 (O145) & constant amplitude, $\beta_2=0$ & 12 & 3.238 & 3.344 & $+0.011$\\
O166 (O152) & gate $1-\tau_{W\sigma}/\tau_\sigma$ & 12 & 3.315 & 3.500 & $+0.156$\\
O189 (O152) & gradient-corrected radius $r_{s\sigma}/(1+u_\sigma)$ & 12 & 3.241 & 3.470 & $+0.126$\\
O163 (O152) & kinetic radius $r_{s\sigma}t_\sigma^{1/5}$ & 12 & 3.282 & 3.362 & $+0.018$\\
O164 (O145) & kinetic radius $r_{s\sigma}t_\sigma^{1/5}$ & 13 & 3.265 & 3.343 & $+0.010$\\
O169 (O125) & exact uniform-gas same-spin limit imposed & 13 & 3.397 & 3.626 & $+0.290$\\
O170 (O145) & exact uniform-gas same-spin limit imposed & 13 & 3.928 & 3.908 & $+0.575$\\
\midrule
O188 (O145) & O145 form from published parent coefficients & 13 & 3.220 & 3.386 & $+0.053$\\
O187 (O178) & O145 form from the O178 optimum, small window & 13 & 3.237 & 3.425 & $+0.092^a$\\
O191 (O152) & O152 form from published parent coefficients & 12 & 3.227 & 3.392 & $+0.047$\\
\midrule
O1 & \wb\ form, training refit (seed) & 13 & 3.184 & 3.870 & ---\\
O27 (O1) & window $h_2$, \wb\ base & 15 & 3.049 & 3.648 & $-0.223^b$\\
O60 (O27) & window $h_2$ with $w_\sigma^2$ amplitude, \wb\ base & 16 & 3.040 & 3.603 & $-0.267^b$\\
O98 (O60) & profile $2r^2/(1+r^4)$, \wb\ base & 16 & 3.074 & 3.543 & $-0.327^b$\\
O160 & B97M-V, uniform-gas constants fixed (control) & 12 & 6.178 & 6.189 & ---\\
O161 (O3) & B97M-V with the same-spin $h_3$ window & 14 & 6.167 & 6.154 & $-0.036^c$\\
\bottomrule\end{tabular}
\par\smallskip\begin{minipage}{0.97\linewidth}\footnotesize
$^a$Relative to O145; relative to the no-window O178 the gain is 0.056. $^b$Relative to the \wb\ seed O1. $^c$Relative to the control O160; both O160 and O161 derive from the B97M-V seed O3. The O185 and O186 changes are relative to O178; relative to O145 they are $+0.349$ and $+0.236$. Values are rounded from the archived records.
\end{minipage}
\end{table}

On this scale, O114, O138, O173, O184, and O186 are indicative, and O137, O150, O152, O161, O163, and O164 are indistinguishable from zero. Because $h_p(r/r_0)$ peaks at $(p-1)^{-1/p}r_0$, the scale $r_0$ and the peak position differ; the peaks of the O125, O137, and O138 windows lie at 0.79, 0.60, and 1.03 bohr, and that of O159 at 0.76 bohr. The GAS22-type same-spin polynomial of O145 and O159 satisfies $P_s(0,0)=0$, so these functionals have no same-spin correlation for the uniform gas, and the window cannot change that limit. O169 and O170 restore the Stoll value $g_{ss}=1$ there and test the base polynomial rather than the window. The magnitude of the fitted window follows from Eq.~\ref{eq:window}: $h_4$ peaks at 0.57 for $r_{s\sigma}=3^{-1/4}\approx0.76$ bohr, so at the peak the multiplier is $1+0.57\beta_0\approx0.85$ for $w_\sigma=0$ and $1+0.57(\beta_0+\beta_2)\approx0.70$ for $|w_\sigma|=1$.

\subsection{Design rule}\label{sec:si-design-rule}
Table~\ref{tab:window-controls} classifies the controls by the robustness scale above.

\begin{table}[htbp]
\centering\small
\setlength{\tabcolsep}{3pt}
\caption{Classification of the refitted controls of the same-spin density window. $\Delta J_{\rm val}$ is the change in validation WRMSD (\kcal) relative to the parent record; each row is a separate joint training fit recorded during the search. Peak positions refer to $r_{s\sigma}$ in bohr.}\label{tab:window-controls}
\begin{tabular}{p{0.35\linewidth}p{0.17\linewidth}p{0.10\linewidth}p{0.29\linewidth}}
\toprule
Change & Record (parent) & $\Delta J_{\rm val}$ & Inference \\
\midrule
\multicolumn{4}{l}{\textit{Robust ($>0.09$ \kcal)}}\\
Window removed & O178 (O145) & $+0.148^a$ & 0.148 of O145's 0.200 gain over \wb \\
Analogous window ($r_s$, $\bar w$) on opposite-spin channel only & O185 (O178) & $+0.201^b$ & specific to same-spin correlation \\
Window on the uniform-gas same-spin term only (\wb\ base) & O136 (O125) & $+0.330$ & must scale the whole same-spin factor \\
Gate $1-\tau_{W\sigma}/\tau_\sigma$ (off in one-orbital regions) & O166 (O152) & $+0.156$ & needed where $\tau_{W\sigma}/\tau_\sigma$ is large \\
Gradient-corrected radius $r_{s\sigma}/(1+u_\sigma)$ & O189 (O152) & $+0.126$ & shifting the window to lower density at large gradient is harmful \\
\multicolumn{4}{l}{\textit{Indicative (0.05--0.09 \kcal)}}\\
One shared window on both channels & O186 (O178) & $+0.088^b$ & not a generic correlation damping \\
Broad $r/(1+r^2)$ & O184 (O145) & $+0.087$ & weight near $r_{s\sigma}\approx2$ costly$^c$ \\
$re^{-r}$ & O173 (O145) & $+0.058$ & as above \\
$h_3(r/1.3)$, peak 0.79 $\to$ 1.03 & O138 (O125) & $+0.083$ & as above \\
Total-density $r_s$ instead of $r_{s\sigma}$ & O114 (O92) & $+0.074$ & shift to lower spin density$^d$ \\
\multicolumn{4}{l}{\textit{Within noise ($<0.05$ \kcal)}}\\
$h_3\to h_4$ & O159 (O145) & $-0.001$ & exponent indistinguishable \\
$2r^2/(1+r^4)$, peak 1.0 & O150 (O145) & $+0.034$ & peak within 0.6--1.0 indistinguishable \\
$h_3(r/0.75)$, peak 0.79 $\to$ 0.60 & O137 (O125) & $+0.015$ & as above \\
Constant amplitude ($\beta_2=0$) & O152 (O145) & $+0.011$ & $w_\sigma$-dependence unresolved \\
Kinetic radius $r_{s\sigma}t_\sigma^{1/5}$ & O164 (O145) & $+0.010$ & indistinguishable from $r_{s\sigma}$ \\
\multicolumn{4}{l}{\textit{Base same-spin factor}}\\
Stoll uniform-gas value restored & O169 (O125); O170 (O145) & $+0.290$; $+0.575$ & base factor far below the Stoll value \\
\bottomrule
\end{tabular}
\par\smallskip\noindent\begin{minipage}{0.97\linewidth}\footnotesize\raggedright $^a$From the archived start; from the no-window optimum (O187) the gain is 0.056 \kcal, which is indicative only. $^b$Relative to O178; both are worse than omitting the window. Both lower the training error more than the same-spin window of O145 does (3.172 and 3.181 versus 3.226 \kcal). $^c$The costly profiles retain about 63\% or more of their peak value at $r_{s\sigma}=2$ bohr; the neutral profiles retain at most 47\%. $^d$For closed shells, $r_s=2^{-1/3}r_{s\sigma}$. O92 and O125 are earlier window lineages built on the GAS22 and \wb\ base forms. A trainable peak scale (O96, parent O92) remained at its initial value. Complete values are in Table~\ref{tab:si-window-controls}.\end{minipage}
\end{table}

The robust controls fix the channel and scope of the window. Relative to the same-spin window of O145, moving the window to the opposite-spin channel (O185, which also uses total-density $r_s$) or sharing it between channels (O186) costs 0.35 and 0.24 \kcal; both lower the training error more than O145 does, so they overfit. A window on the uniform-gas same-spin term alone, a gate that switches the window off where $\tau_{W\sigma}/\tau_\sigma$ is large, and a gradient-corrected length scale all cost more than 0.12 \kcal. Together, these controls define an empirical design rule for this range-separated hybrid: reduce the magnitude of the entire same-spin enhancement factor at intermediate spin density, without a $\tau_W/\tau$ gate, a radius that moves it to lower density, or an opposite-spin counterpart. The controls come from three lineages (O92, O125, and O145), and the whole-factor inference (O136) was obtained on the \wb\ base. The rule applies to base factors of the GAS22 type, which vanish for the uniform gas and lie far below the Stoll value. O189 is consistent with the low-density result below: its radius moves the $h_3$ window peak to $r_{s\sigma}\approx0.79(1+u_\sigma)$, toward 1.6 bohr in high-gradient regions.

The fitted amplitude is consistent across forms: a constant-amplitude window converged to $\beta_0=-0.39$ and $-0.37$ from the two tested starts of the $h_3$ form (O152, O191) and to $-0.32$ with $h_4$ (O197, validation 3.361 \kcal), reductions of 18--21\% at the window's peak, and the $w_\sigma$-dependent amplitude of O159 gives 15--30\%; repeated refits of O159 (Section~\ref{sec:si-attenuation}) reach two basins with $\beta_2=-0.27$ or $-0.36$ at validation errors within 0.006 \kcal, so the $w_\sigma$ dependence is unresolved.

The indicative controls add that the window should not remain near its peak value at $r_{s\sigma}=2$ bohr ($\rho_\sigma\approx0.03$ a.u.): profiles retaining about 63\% or more of their peak there cost 0.06--0.09 \kcal, whereas narrow profiles, retaining at most 47\%, peaking anywhere between 0.6 and 1.0 bohr cannot be distinguished (a threshold identified post hoc from these five profiles). The O159 window has a half-maximum range of $r_{s\sigma}\approx0.3$--1.4 bohr ($\rho_\sigma\approx9$--0.09 a.u.). Restoring the Stoll uniform-gas value of the base factor costs 0.29--0.58 \kcal. The sign of the window energy is analyzed below.

Analogous windows also improve the \wb\ base. Windows of the form $h_2$ or $2r^2/(1+r^4)$ lower the validation error of the overfitted \wb\ training refit (training 3.184, validation 3.870 \kcal) to 3.543--3.648 \kcal\ (O27, O60, O98), close to published \wb\ (3.533 \kcal). In the nonhybrid B97M-V form, the change (6.189 to 6.154 \kcal; O160, O161) is within noise.

\subsection{Sign of the window energy}
The GAS22-type same-spin polynomial changes sign with $w_\sigma$. For $w_\sigma\gtrsim0.3$, $g_{ss}e_{c\sigma\sigma}^{\rm PW}$ is positive in the fitted O159 functional, and a multiplier below one lowers the energy; for $w_\sigma\lesssim0$ it is predominantly negative, and the multiplier raises the energy. Near $w_\sigma=0$ the sign depends on $u_\sigma$. Summed over all validation molecules, the window energy is $-5.9\times10^4$ \kcal\ from $w_\sigma\geq0.3$ and $+8.8\times10^4$ \kcal\ from $w_\sigma\leq-0.3$. The window therefore reduces the magnitude of the same-spin factor; it does not uniformly remove correlation energy.

\subsection{Error decomposition against the refitted control}
Table~\ref{tab:si-window-decomposition} decomposes the O178-to-O159 change in squared WRMSD by category. The contributions are exact and sum to the total; negative values reduce $J^2$. Subset values are root-mean-square deviations of the subset; category and subset shares are computed with the same exact decomposition as the descriptor bins below.

Relative to the refitted control, O159 lowers the squared validation WRMSD mainly through barrier heights (35\% of the reduction), easy thermochemistry (19\%), difficult isomerizations (14\%), and noncovalent clusters (12\%); a further 22\% is attributed to the heavily weighted RG10 (Table~\ref{tab:si-window-decomposition}, note a). The largest subset gains are for Shields38, HTBH38 (1.71 to 1.49 \kcal), HAT707nonMR, WATER27, ISOMERIZATION20, and CR20, whereas the sulfate--water anion clusters of SW49 and the electron affinities of EA13 worsen. On training data, 62\% of the reduction comes from the AE18 atomic energies (RMSD 7.60 to 5.56 \kcal; Section~\ref{sec:si-ae18}).

On the held-out MGCDB84 test split, O159 lowers the non-SCF WRMSD relative to O178 from 3.865 to 3.818 \kcal, about one third of the validation difference; the direction of individual categories is more informative than this net change. Noncovalent clusters ($-0.487$ in the squared WRMSD), easy isomerizations ($-0.357$), and barrier heights ($-0.239$) improve; as on validation, water clusters and hydrogen- and proton-transfer barriers improve (H2O20Bind4, 1.26 to 0.69 \kcal; WCPT27, 1.86 to 1.70 \kcal; PX13), and ion-containing binding worsens (Ionic43, 0.61 to 0.67 \kcal). Difficult isomerizations reverse, improving on validation but worsening on test ($+0.423$, dominated by the eight C20C24 reactions), and easy noncovalent dimers worsen on both splits ($+0.260$ on test). Most category directions found on validation thus recur on held-out data.

\begin{table}[htbp]\centering\small
\caption{O178 (refitted, no window) to O159: category contributions to $\Delta J^2$ (\kcal$^2$; negative values are improvements) and selected subset RMSDs (\kcal).}\label{tab:si-window-decomposition}
\begin{tabular}{p{0.16\linewidth}p{0.36\linewidth}p{0.40\linewidth}}\toprule
Split, $J$ & Category contributions to $\Delta J^2$ & Subsets, O178 $\to$ O159\\\midrule
Validation, $3.481\to3.332$ ($\Delta J^2=-1.015$) & BH $-0.358$, RG10 $-0.228^a$, TCE $-0.194$, ID $-0.144$, NCEC $-0.122$, TCD $-0.023$, IE $-0.022$, NCD $-0.009$, NCED $+0.086$ & improved: Shields38 0.49$\to$0.35, HTBH38 1.71$\to$1.49, HAT707nonMR 2.67$\to$2.59, WATER27 0.49$\to$0.42, ISOMERIZATION20 2.32$\to$2.18, CR20 0.81$\to$0.57; worsened: SW49Bind6 0.55$\to$0.70, SW49Bind345 0.25$\to$0.32, EA13 1.78$\to$1.92, BzDC215, S66x8\\
Training, $3.386\to3.252$ ($\Delta J^2=-0.890$) & AE18 $-0.554$, ID $-0.291$, BH $-0.148$, NCED $-0.023$, NCD $-0.009$, TCE $0.000$, NCEC $+0.041$, IE $+0.095$ & improved: AE18 7.60$\to$5.56, Styrene45 2.03$\to$1.89, BHPERI26 1.24$\to$1.02, DBH24 1.58$\to$1.46; worsened: CRBH20 1.02$\to$1.37, G21EA 2.48$\to$2.66\\
Test, $3.865\to3.818$ ($\Delta J^2=-0.361$) & NCEC $-0.487$, IE $-0.357$, BH $-0.239$, TCE $+0.010$, TCD $+0.028$, NCED $+0.260$, ID $+0.423$ & improved: H2O20Bind4 1.26$\to$0.69, WCPT27 1.86$\to$1.70, PX13 1.92$\to$1.82, YMPJ519, H2O20Bind10; worsened: C20C24 5.39$\to$5.72 (8 reactions), Ionic43 0.61$\to$0.67, BSR36\\
\bottomrule\end{tabular}
\par\smallskip\begin{minipage}{0.97\linewidth}\footnotesize
$^a$The RG10 contribution arises from its weight; its RMSD changes only from 0.183 to 0.182 \kcal. Test values use full-grid non-SCF energies. The test split was not used for selection, but the descriptor analysis was performed after the test results were known. The net test change (0.047 \kcal\ in $J$) is within the initialization spread of the fits.
\end{minipage}
\end{table}

\subsection{Atomic energies}\label{sec:si-ae18}
Table~\ref{tab:si-ae18} gives AE18 total-energy errors for the refitted control and O159, together with the window energy $E_{\rm win}$ of each atom. AE18 is part of the training set. The control's total energies show a period-dependent error: they are too low from C to Si and too high from P to Ar. O159 reduces both errors, except for Si, which is unchanged, whereas the errors for H to B grow. Because O178 and O159 are separate refits, the change is not the window energy alone. For C to Si, $E_{\rm win}$ is positive and has the sign that reduces the control's error; for P to Ar, $E_{\rm win}$ raises energies that are already too high, so the improvement there comes from the co-adapted coefficients. $E_{\rm win}$ increases with atomic number overall but not monotonically (for example, O is lower than N). The improvement is therefore not specific to high-spin atoms: P changes only from 5.48 to 4.48 \kcal. Because the window energy grows from $-2.6$ \kcal\ (He) to $+10.8$ \kcal\ (Ar), part of the AE18 training gain reflects a $Z$-dependent correction of atomic total energies, which reappears for heavier atoms in Section~\ref{sec:si-window-control}. For second-period atoms, most of the positive window energy lies at $r_{s\sigma}\approx0.4$--0.8 bohr (N, $+4.5$ \kcal; O, $+4.9$ \kcal), partly offset by the core region $r_{s\sigma}\lesssim0.4$ bohr. For Na to S, the positive contribution moves to $r_{s\sigma}\approx0.8$--1.5 bohr. Because the window raises the energy only where the same-spin term is negative, which for the GAS22-type polynomial is mainly at $w_\sigma\lesssim0$, the atomic window energies arise largely outside the $w_\sigma\geq0.3$ region that dominates the reaction-energy attribution below.

\begin{table}[htbp]\centering\small
\caption{AE18 total-energy errors (calculated minus reference, \kcal) for the refitted no-window control O178 and for O159, and the O159 window energy $E_{\rm win}$ (\kcal).}\label{tab:si-ae18}
\begin{tabular}{lrrr@{\hspace{2em}}lrrr}\toprule
Atom & O178 & O159 & $E_{\rm win}$ & Atom & O178 & O159 & $E_{\rm win}$\\\midrule
H & 2.25 & 2.98 & $-0.83$ & Ne & $-9.90$ & $-7.11$ & 3.54\\
He & 0.94 & 1.65 & $-2.62$ & Na & $-7.44$ & $-6.25$ & 3.69\\
Li & 0.62 & 2.51 & $-2.07$ & Mg & $-10.66$ & $-9.79$ & 4.54\\
Be & $-0.30$ & 4.22 & $-0.32$ & Al & $-6.85$ & $-6.26$ & 5.62\\
B & $-1.74$ & 4.81 & 0.66 & Si & $-1.74$ & $-1.74$ & 6.81\\
C & $-5.38$ & 1.54 & 1.72 & P & 5.48 & 4.48 & 8.22\\
N & $-7.02$ & $-0.56$ & 3.20 & S & 6.13 & 4.67 & 8.89\\
O & $-12.17$ & $-6.81$ & 1.63 & Cl & 8.14 & 5.25 & 9.73\\
F & $-13.54$ & $-9.26$ & 2.19 & Ar & 13.00 & 8.15 & 10.79\\
\bottomrule\end{tabular}
\end{table}

\subsection{Descriptor-resolved attribution}
For each molecule, the window energy $E_{\rm win}$ was binned on 30 logarithmic $r_{s\sigma}$ intervals between 0.05 and 20 bohr, with underflow and overflow bins, and 20 intervals of $w_\sigma$, summing over grid points and spin channels. The bins reproduce each molecular window energy to within about $2\times10^{-11}\Eh$. Three models were compared: O159 (``on''), O159 with $\beta_0=\beta_2=0$ and all other coefficients unchanged (``off''), and the refitted control O178. The split WRMSDs are listed in Table~\ref{tab:si-window-localization}. With reaction weights $w_i$, signed off-model errors $e_i$, and bin contributions $E_{{\rm win},ib}$ to the reaction window energy $E_{{\rm win},i}$,
\begin{equation}
J_{\rm on}^2-J_{\rm off}^2=\sum_b\frac1{N_S}\sum_{i\in S} w_i(2e_i+E_{{\rm win},i})E_{{\rm win},ib}
\end{equation}
is an exact attribution of the change to descriptor bins. Neighboring bins carry large contributions of opposite sign, so only regional totals are interpreted. Because $h_4$ confines the multiplier to roughly $0.3<r_{s\sigma}<1.4$ bohr by construction, the $r_{s\sigma}$ shares are expected; the informative result is the $w_\sigma$ localization. By category, 98\% of the HTBH38 contribution arises at $0.82<r_{s\sigma}<1.49$ bohr, and the Shields38 and WATER27 contributions lie mostly at $0.37<r_{s\sigma}<1.49$ bohr with $w_\sigma\geq0.3$. The $w_\sigma$ bins have width 0.1, so $w_\sigma\geq0.3$ corresponds to $\tau_\sigma\leq0.54\,\tau_{U\sigma}$; the region $0.37<r_{s\sigma}<1.49$ bohr with $w_\sigma\geq0.4$ accounts for 75\% (validation) and 88\% (test) of the change, from 3.1\% and 3.2\% of the electrons. The harmful SW49Bind contributions lie mostly (about 70\%) in the same region ($0.8<r_{s\sigma}<1.5$ bohr, $w_\sigma\geq0.3$) as the gains; the tail $r_{s\sigma}>3$ bohr contributes slightly in the opposite direction.

Attribution at fixed coefficients shows where the window matters for reaction energies, as distinct from where it acts: the large window energies at $w_\sigma\leq-0.3$ hardly affect the reaction errors. This decomposition partitions the fixed-coefficient effect, which is about 20 (validation) and 100 (test) times the refitted gain over O178 in $J^2$; it therefore maps where the other O159 coefficients depend on the window. Since $\tau_\sigma/\tau_{U\sigma}=\alpha_\sigma+5s_\sigma^2/3$, with $\alpha_\sigma=(\tau_\sigma-\tau_{W\sigma})/\tau_{U\sigma}$ and $s_\sigma$ the reduced gradient, the region $w_\sigma\geq0.3$ requires $\alpha_\sigma+5s_\sigma^2/3\lesssim0.54$: low-gradient, single-orbital or bonding regions, such as those near protons and at bond midpoints, rather than single-orbital density tails, which lie at $w_\sigma\to-1$.

\begin{table}[htbp]\centering\small
\caption{Window removal and localization of its effect. WRMSDs in \kcal; shares are fractions of $J_{\rm on}^2-J_{\rm off}^2$.}\label{tab:si-window-localization}
\begin{tabular}{lrrrrrrr}\toprule
Split & $J_{\rm on}$ & $J_{\rm off}$ & $J_{\rm O178}$ & $0.37<r_{s\sigma}<1.49$ & $w_\sigma\geq0.3$ & Both & Electrons in both\\\midrule
Training & 3.252 & 5.111 & 3.386 & --- & --- & --- & ---\\
Validation & 3.332 & 5.595 & 3.481 & 87\% & 82\% & 82\% & 4.9\%\\
Test & 3.818 & 7.188 & 3.865 & 88\% & 91\% & 91\% & 5.4\%\\
\bottomrule\end{tabular}
\end{table}

\subsection{Same-spin self-correlation of the hydrogen atom}
The exact hydrogen-atom density was used to evaluate the semilocal correlation energy (VV10 excluded), which should vanish for a one-electron system (Table~\ref{tab:si-hatom}). The window lowers O159's error relative to the same coefficients without the window, but the refitted control without the window has a smaller error than O159. The window gain is therefore not a correction of one-electron self-correlation. Along the hydrogen-atom density, the distance $r=0$, 0.25, 0.5, and 1.0 bohr maps to $(r_{s\sigma},w_\sigma)=(0.91,0.62)$, $(1.07,0.51)$, $(1.27,0.37)$, and $(1.77,0.06)$, so the region within 0.5 bohr of the proton lies inside the band identified above. Within O159, the window acts as a partial, density-localized reduction of spurious anti-correlating same-spin energy in one-orbital-like regions, and it is needed where $\tau_{W\sigma}/\tau_\sigma$ is large (O166). The controls also argue against a channel-independent damping of correlation (O185, O186).

\begin{table}[htbp]\centering\small
\caption{Semilocal correlation energy of the exact hydrogen-atom density (\kcal; exact value 0).}\label{tab:si-hatom}
\begin{tabular}{lr}\toprule
Functional & $E_c$(H)\\\midrule
\wb\ (published) & $+4.11$\\
GAS22 (published) & $+4.36$\\
O178 (refitted, no window) & $+2.50$\\
O145 & $+4.14$\\
O159 & $+3.52$\\
O159, window removed at fixed coefficients & $+4.55$\\
\bottomrule\end{tabular}
\end{table}

\subsection{Interpretation}
A simple interpretation accounts for these results. The same-spin term inherits its density dependence from the same-spin PW92 energy, while its B97-type enhancement factor depends only on $w_\sigma$ and $u_\sigma$, which are invariant under uniform coordinate scaling. At fixed $w_\sigma$ and $u_\sigma$, the factor therefore cannot be reduced in one density range without being changed in others, and the fit leaves it too large in magnitude, and of anti-correlating sign, in low-$\alpha_\sigma$ regions of intermediate spin density. A multiplicative window in $r_{s\sigma}$ that tends to unity at both density limits adds an independent density scale without altering either limit; accordingly, a kinetic radius performs as well (O164). The closest analogue in nonempirical functionals is the $\alpha=0$ correlation channel of SCAN-type functionals, which carries its own $r_s$ dependence; the window retrofits a density-dependent, low-$\alpha$ correction onto a scale-invariant B97 same-spin series. Multiplicative $r_s$ corrections of B97-type factors are not themselves new: SAFS26-b applies broad corrections of the form $1+\beta r_s/(1+0.1r_s^2)$ to both correlation channels. What the present controls add is their characterization in this frame: the window should act on same-spin correlation only, scale the whole factor, carry no $\tau_W/\tau$ gate, and be small at $r_{s\sigma}\approx2$ bohr; the broad two-channel SAFS26-b form satisfies neither the first nor the last condition. The characterization makes a falsifiable prediction: any length variable that keeps the window peaked at $r_{s\sigma}\approx0.6$--1.0 bohr and small near 2 bohr should perform equivalently, and an opposite-spin or broad window should again overfit.

\subsection{External comparison with \wb}
Table~\ref{tab:si-window-gscdb} lists all 20 GSCDB137 datasets without MGCDB84 antecedents, with O159 SCF and local \wb\ SCF results. O159 lowers the NER for 10 sets and raises it for 10. AE11, O24, and O24x4 improve and P34EA worsens; the transition-metal, field-response, and vibrational sets also worsen. Section~\ref{sec:si-window-control} separates the window from the refit of the other coefficients by comparison with O178.

\begin{table}[htbp]\centering\small
\caption{GSCDB137 datasets without MGCDB84 antecedents: benchmark error measure and NER for local \wb\ SCF and O159 SCF.}\label{tab:si-window-gscdb}
\begin{tabular}{llrrrr}\toprule
Dataset & Type & \wb & O159 & NER \wb & NER O159\\\midrule
3d4dIPSS & TM & 5.073 & 5.613 & 0.977 & 1.081\\
AE11 & TC & 368.86 & 295.66 & 3.154 & 2.528\\
ALKBDE10 & TC & 3.581 & 3.561 & 1.100 & 1.094\\
DAPD & TM & 6.264 & 6.625 & 2.256 & 2.386\\
Dip146 & EF & 0.0365 & 0.0391 & 1.051 & 1.125\\
DIPCS9 & TC & 5.106 & 4.189 & 1.510 & 1.238\\
HCP32 & NC & 0.503 & 0.515 & 0.879 & 0.901\\
He3 & NC & 0.183 & 0.172 & 1.038 & 0.973\\
HEAVY28 & NC & 0.178 & 0.155 & 0.943 & 0.821\\
IP23 & TC & 2.519 & 2.461 & 0.991 & 0.969\\
MB08-165 & TC & 3.791 & 3.986 & 0.829 & 0.872\\
O24 & NC & 1.028 & 0.903 & 0.914 & 0.803\\
O24x4 & NC & 0.603 & 0.580 & 0.860 & 0.828\\
OEEF & EF & 0.0993 & 0.103 & 1.134 & 1.175\\
ORBH35 & BH & 2.584 & 2.544 & 1.140 & 1.122\\
P34EA & TC & 2.791 & 2.997 & 1.777 & 1.909\\
Pol130 & EF & 0.0478 & 0.0490 & 2.010 & 2.062\\
SIE4x4 & TC & 10.61 & 10.37 & 1.156 & 1.130\\
TMD10 & TM & 6.458 & 7.003 & 1.144 & 1.241\\
V30 & FQ & 27.83 & 33.55 & 0.926 & 1.116\\
\bottomrule\end{tabular}
\par\smallskip\begin{minipage}{0.97\linewidth}\footnotesize
Types: TM, transition metal; TC, thermochemistry; NC, noncovalent; BH, barrier height; EF, electric field; FQ, frequency. The error measure is the benchmark-specified metric of each dataset (mean absolute error in \kcal\ for AE11); NER divides it by the fixed published normalization.
\end{minipage}
\end{table}
\FloatBarrier

\subsection{External test against the refitted control}\label{sec:si-window-control}
The refitted no-window control O178 was evaluated non-self-consistently on the 36 GSCDB137 datasets at the same stored \wb\ densities as O159. O159 and O178 share the exact-exchange and VV10 contributions, so
\begin{equation}
E^{\rm nonSCF}[\mathrm{O178}]=E^{\rm nonSCF}[\mathrm{O159}]+E_{\rm sl}[\mathrm{O178};\rho^0]-E_{\rm sl}[\mathrm{O159};\rho^0],
\end{equation}
where the semilocal difference is obtained by integrating the difference of the two semilocal energy densities on the production grid of each calculation. O159 energies are taken at full precision from the archived molecular records. The shortcut agrees with a direct O178 non-SCF evaluation to within $3\times10^{-14}\Eh$ for five molecules covering closed-shell, open-shell, field-dependent, transition-metal, and one-electron cases, and the recomputed O159 NERs reproduce the archived values exactly. As for O159, the archived attenuation kernel was used (Section~\ref{sec:si-attenuation}), and the Rn atom of AE11 was treated as in the original calculation. The difference calculation required about 0.5 GPU-h.

Before the O178 energies were computed, a direction was recorded, with a timestamp preserved in the archive, for each dataset with a corresponding MGCDB84 training, validation, or test trend. For the new datasets, these were improvement for AE11 (extrapolating the AE18 trend), IP23 (IP13 improved), and HCP32 (XB51 and XB18 improved); worsening for P34EA (EA13 and G21EA worsened) and for O24 and O24x4 (noncovalent dimers worsened slightly); and no change for He3, whose energies are very small. Confidence was recorded as low except for P34EA and He3 (medium). For datasets with MGCDB84 antecedents, the prediction was intended to follow the sign of the corresponding MGCDB84 subset change against O178, or neutral where that change was negligible. Two recorded predictions departed from this rule: WCPT6 was predicted to improve from the WCPT27 test gain, although the MGCDB84 test subset WCPT6 itself worsened (RMSD 0.414 to 0.584 \kcal), and CT20 was recorded as neutral despite a validation gain of 2.6\% (0.106 to 0.103 \kcal). The predictions are scored as recorded. Transition-metal, field-response, and vibrational datasets, and ALKBDE10, DIPCS9, MB08-165, SIE4x4, HEAVY28, and ORBH35, received no prediction. A dataset is neutral if its error measure changes by less than 1\% relative to O178; a directional prediction is correct if the sign matches and the dataset is not neutral, and a neutral prediction is correct if the dataset is neutral. The primary aggregate, specified with the predictions, is the mean NER over the 20 new datasets; the other scopes are secondary. The aggregate statistics use datasets as sampling units (paired signed-rank test; bootstrap with 20,000 resamples). The expected aggregate effect, recorded with the predictions, was small or null on the new datasets, because the MGCDB84 test gain against O178 is only 0.047 \kcal. The predictions were recorded after the O159 and \wb\ GSCDB137 results were known.

Table~\ref{tab:si-window-control} lists every dataset. On the new datasets, 3 of 7 scored predictions were correct, and on the datasets with antecedents 11 of 16. Predicting improvement for every dataset would have scored 12 of 16 and 5 of 7. The window improves 26 of 36 datasets, but the aggregate gain on the new datasets is carried by AE11. Among datasets without a prediction, the largest changes are losses for V30 ($+29.6$\%) and DIPCS9 ($+22.1$\%) and gains for MB08-165 ($-15.9$\%) and OEEF ($-14.1$\%). Comparing the O178, O159, and \wb\ NER columns separates two sources of O159's differences from \wb. For the transition-metal datasets that O159 describes less accurately than \wb\ (3d4dIPSS, DAPD, TMD10), O178 is less accurate still, so these losses arise from the refitted base form rather than from the window. For V30 and Pol130, O178 is at least as accurate as \wb, so these losses accompany the window. Within the antecedent group, G21EA improves by 20\% although the MGCDB84 G21EA subset worsened against O178 (2.48 to 2.66 \kcal). WCPT6 worsens by 24.5\%, as its MGCDB84 test subset did, contrary to the recorded prediction; applying the stated rule to WCPT6 and CT20 would have given 13 of 16 correct. The antecedent datasets overlap the MGCDB84 selection data, and AE18 is part of the training set, so agreement on them tests reproducibility under revised references.

\begin{table}[htbp]\centering\small
\setlength{\tabcolsep}{3.5pt}
\caption{GSCDB137 per-dataset comparison of O159 with the refitted no-window control O178 (both non-SCF at the stored \wb\ densities). Error measures are the benchmark-specified metrics; $\delta$ is the relative change of the error measure from O178 to O159 (\%). The \wb\ NER is the local SCF reference.}\label{tab:si-window-control}
\begin{tabular}{llrrrrrrl}\toprule
Dataset & Type & O178 & O159 & $\delta$ & NER O178 & NER O159 & NER \wb & Prediction\\\midrule
\multicolumn{9}{l}{\textit{Datasets without MGCDB84 antecedents}}\\
3d4dIPSS & TM & 5.717 & 5.616 & $-1.8$ & 1.101 & 1.081 & 0.977 & --- \\
AE11 & TC & 415.8 & 295.8 & $-28.9$ & 3.555 & 2.529 & 3.154 & better (l) \checkmark\\
ALKBDE10 & TC & 3.703 & 3.568 & $-3.6$ & 1.138 & 1.096 & 1.100 & --- \\
DAPD & TM & 7.054 & 6.632 & $-6.0$ & 2.540 & 2.388 & 2.256 & --- \\
Dip146 & EF & 0.0375 & 0.0392 & $+4.6$ & 1.079 & 1.128 & 1.051 & --- \\
DIPCS9 & TC & 3.426 & 4.183 & $+22.1$ & 1.013 & 1.237 & 1.510 & --- \\
HCP32 & NC & 0.443 & 0.514 & $+16.1$ & 0.775 & 0.899 & 0.879 & better (l) $\times$\\
He3 & NC & 0.178 & 0.171 & $-3.9$ & 1.013 & 0.973 & 1.038 & neutral (m) $\times$\\
HEAVY28 & NC & 0.164 & 0.155 & $-5.0$ & 0.865 & 0.822 & 0.943 & --- \\
IP23 & TC & 2.487 & 2.460 & $-1.1$ & 0.979 & 0.968 & 0.991 & better (l) \checkmark\\
MB08-165 & TC & 4.769 & 4.008 & $-15.9$ & 1.043 & 0.876 & 0.829 & --- \\
O24 & NC & 0.983 & 0.908 & $-7.6$ & 0.874 & 0.808 & 0.914 & worse (l) $\times$\\
O24x4 & NC & 0.590 & 0.583 & $-1.2$ & 0.841 & 0.831 & 0.860 & worse (l) $\times$\\
OEEF & EF & 0.120 & 0.103 & $-14.1$ & 1.367 & 1.174 & 1.134 & --- \\
ORBH35 & BH & 2.616 & 2.543 & $-2.8$ & 1.155 & 1.122 & 1.140 & --- \\
P34EA & TC & 2.524 & 3.007 & $+19.1$ & 1.607 & 1.915 & 1.777 & worse (m) \checkmark\\
Pol130 & EF & 0.0476 & 0.0491 & $+3.2$ & 2.003 & 2.068 & 2.010 & --- \\
SIE4x4 & TC & 10.509 & 10.368 & $-1.3$ & 1.145 & 1.130 & 1.156 & --- \\
TMD10 & TM & 7.210 & 7.016 & $-2.7$ & 1.278 & 1.243 & 1.144 & --- \\
V30 & FQ & 25.902 & 33.573 & $+29.6$ & 0.862 & 1.117 & 0.926 & --- \\
\midrule
\multicolumn{9}{l}{\textit{Datasets with MGCDB84 antecedents}}\\
AE18 & TC & 6.327 & 4.925 & $-22.2$ & 1.278 & 0.995 & 1.072 & better \checkmark\\
BDE99MR & TC & 4.135 & 4.222 & $+2.1$ & 1.248 & 1.274 & 1.228 & neutral $\times$\\
BDE99nonMR & TC & 1.606 & 1.556 & $-3.1$ & 0.832 & 0.806 & 0.790 & better \checkmark\\
BH76RC & TC & 0.974 & 0.902 & $-7.4$ & 0.968 & 0.897 & 0.846 & better \checkmark\\
CT20 & NC & 0.0713 & 0.0687 & $-3.7$ & 0.797 & 0.768 & 0.818 & neutral $\times$\\
DBH22 & BH & 1.227 & 1.151 & $-6.2$ & 1.017 & 0.954 & 0.973 & better \checkmark\\
G21EA & TC & 1.681 & 1.343 & $-20.1$ & 1.228 & 0.982 & 1.001 & worse $\times$\\
HAT707MR & TC & 3.629 & 3.578 & $-1.4$ & 1.130 & 1.114 & 1.090 & better \checkmark\\
HAT707nonMR & TC & 1.935 & 1.818 & $-6.1$ & 0.819 & 0.769 & 0.805 & better \checkmark\\
ISOMERIZATION20 & IS & 1.824 & 1.734 & $-4.9$ & 1.394 & 1.326 & 1.166 & better \checkmark\\
NC11 & NC & 0.0420 & 0.0413 & $-1.8$ & 0.773 & 0.759 & 0.829 & neutral $\times$\\
RG10N & NC & 0.0840 & 0.0847 & $+0.8$ & 0.790 & 0.797 & 0.872 & neutral \checkmark\\
SN13 & TC & 0.714 & 0.592 & $-17.2$ & 1.132 & 0.937 & 0.799 & better \checkmark\\
TA13 & NC & 1.611 & 1.613 & $+0.1$ & 1.187 & 1.189 & 1.228 & neutral \checkmark\\
TAE\_W4-17MR & TC & 4.423 & 4.350 & $-1.7$ & 1.438 & 1.414 & 1.345 & better \checkmark\\
WCPT6 & TC & 0.364 & 0.453 & $+24.5$ & 0.548 & 0.682 & 0.517 & better $\times$\\
\bottomrule\end{tabular}
\par\smallskip\begin{minipage}{0.97\linewidth}\footnotesize
Types as in Table~\ref{tab:si-window-gscdb}; IS, isomerization. Predictions were recorded before O178 was evaluated; (l) and (m) denote low and medium confidence, which was recorded only for the new datasets. \checkmark, correct; $\times$, incorrect; ---, no prediction. Rounded relative changes near $\pm1$\% are classified from unrounded values.
\end{minipage}
\end{table}
\FloatBarrier

\section{Performance comparisons}\label{sec:si-results}
\subsection{MGCDB84: primary split and selected candidates}
Table~\ref{tab:si-mgc-syfes} compares the completed full-grid calculations on the discovery partition. O159's improvement over \wb\ persists under SCF, while GAS22 retains the lower test error. All 5,930 O159 calculations converged under the recorded protocol. GAS22 retains 20 unconverged molecular records; excluding the 31 affected reactions changes split WRMSDs by at most 0.011 \kcal. Both complete and exclusion-based results are archived.
\begin{table}[htbp]
\centering\small
\setlength{\tabcolsep}{5pt}
\caption{MGCDB84 SyFES: direct full-grid non-SCF and SCF WRMSDs (kcal mol$^{-1}$).}\label{tab:si-mgc-syfes}
\begin{tabular}{llrrrr}
\toprule
Functional & Mode & Train & Validation & Test & VT$_{\neg\mathrm{RG10}}$ \\
\midrule
$\omega$B97M-V & SCF & 3.319 & 3.533 & 4.309 & 3.867 \\
GAS22 & non-SCF & 3.270 & 3.462 & 3.599 & 3.657 \\
GAS22 & SCF & 3.264 & 3.470 & 3.589 & 3.664 \\
O159 & non-SCF & 3.252 & 3.332 & 3.818 & 3.613 \\
O159 & SCF & 3.253 & 3.330 & 3.805 & 3.607 \\
\bottomrule
\end{tabular}
\par\smallskip\noindent\begin{minipage}{0.97\linewidth}\footnotesize\raggedright Published parameters for both reference functionals; frozen fitted parameters for O159. Counts (train/validation/test/VT): 870/2963/1152/3546. VT excludes RG10. GAS22 SCF retains 20 unconverged molecular records; the sensitivity analysis is reported separately.\end{minipage}
\end{table}

The later fixed-density candidates have lower validation errors but higher test errors than O159 (Table~\ref{tab:short-candidates}). C356 improves validation by 3.6\% relative to its training-refitted GAS22 seed but worsens test by 7.4\%. Published and refitted GAS22 are different parameter sets. The other four Codex shortlisted candidates were not tested; no missing SCF or test score is inferred from their eligibility.
\begin{table}[htbp]\centering\small
\caption{Selected cached non-SCF results on the SyFES split (\kcal).}\label{tab:short-candidates}
\begin{tabular}{lrrrr}\toprule
Candidate & Parameters & Training & Validation & Test\\\midrule
O159 &13&3.252&3.332&3.818\\
O396 &10&3.140&3.138&3.841\\
O388 &9&3.191&3.236&3.906\\
O376 &12&3.230&3.294&3.961\\
C356 &14&3.207&3.363&3.897\\
GAS22 refit (Codex) &12&3.266&3.489&3.628\\\bottomrule
\end{tabular}
\par\smallskip\begin{minipage}{0.96\linewidth}\footnotesize
All test values use frozen parameters. Cached scores are distinguished from the full-grid evaluations in Table~\ref{tab:si-mgc-syfes}. Exact values, category errors, and the complete shortlisted-record table are archived.
\end{minipage}
\end{table}

\subsection{Comparison with the SAFS26 reporting partition}
Table~\ref{tab:safs-comparison} brings together all seven functionals reported in SAFS26 and the locally tested comparators for which the required split metrics are available. Its explicit $\mathrm{WRMSD}_{\rm tot}$ column combines validation and test excluding RG10, retaining their existing weights. Panel A contains the original paper's non-SCF results. Panels B/C use the local C20C24-reassigned reporting variant, which approximates that paper's partition; the exact transferred molecule identities and fitted SAFS26 coefficients are unavailable.\cite{duston2026agentic} The 3,547 versus 3,546 combined-observation counts make the difference in scope explicit.

Published-reference and O159 values come from the completed full-grid reevaluation included in the reproducibility archive. Later candidates and the Codex reference refits are reaggregated from archived validation, test, and category squared errors: all eight C20C24 test reactions enter validation with twice their former weight. No candidate is refitted, reselected, or newly evaluated. The archived reaction-weight files verify this transfer rule, and it reproduces all five saved full-grid reaggregations to numerical precision. Seed-only B97M-V and untested shortlist records cannot be transferred without their missing C20C24 test errors and therefore have no comparable row.

On the local reporting variant, combined SCF WRMSDs are 4.090 for \wb, 3.793 for GAS22, and 3.767 for O159. GAS22 still has the lower test error (3.161 versus 3.304 \kcal). Cross-paper differences in the table combine different partitions, parameter sets, and SCF/non-SCF modes.
\begin{table}[p]
\centering\small
\setlength{\tabcolsep}{7pt}
\caption{MGCDB84 comparison with SAFS26: reported and approximately matched splits (kcal mol$^{-1}$).}\label{tab:safs-comparison}
\begin{tabular}{lrrrr}
\toprule
Functional & Training & Validation & Test & $\mathrm{WRMSD}_{\rm tot}$\\
\midrule
\multicolumn{5}{l}{\textit{A. SAFS26 paper: reported non-SCF results}}\\
$\omega$B97M-V & 3.32 & 4.02 & 3.64 & 4.07\\
GAS22 & 3.27 & 3.74 & 3.19 & 3.77\\
SAFS26-x & 2.91 & 3.95 & 3.49 & 3.81\\
SAFS26-a & 2.76 & 3.65 & 3.41 & 3.70\\
SAFS26-b & 2.75 & 3.78 & 3.44 & 3.83\\
AFS26-a & 2.81 & 3.74 & 3.78 & 3.90\\
AFS26-b & 2.65 & 3.65 & 3.96 & 3.90\\
\midrule
\multicolumn{5}{l}{\textit{B. This work: non-SCF and SCF reference}}\\
$\omega$B97M-V (SCF reference) & 3.319 & 4.086 & 3.632 & 4.090\\
GAS22 & 3.270 & 3.778 & 3.168 & 3.788\\
O159 & 3.252 & 3.736 & 3.307 & 3.776\\
O396$^{a}$ & 3.140 & 3.640 & 3.225 & 3.664\\
O388$^{a}$ & 3.191 & 3.796 & 3.196 & 3.786\\
O376$^{a}$ & 3.230 & 3.855 & 3.249 & 3.844\\
C356$^{a}$ & 3.207 & 3.756 & 3.410 & 3.849\\
GAS22 refit (Codex)$^{a}$ & 3.266 & 3.804 & 3.198 & 3.818\\
$\omega$B97M-V refit (Codex)$^{a}$ & 3.191 & 4.363 & 4.052 & 4.368\\
\midrule
\multicolumn{5}{l}{\textit{C. This work: candidate SCF results}}\\
GAS22 & 3.264 & 3.782 & 3.161 & 3.793\\
O159 & 3.253 & 3.726 & 3.304 & 3.767\\
\bottomrule
\end{tabular}
\par\smallskip\begin{minipage}{0.97\linewidth}\footnotesize
$\mathrm{WRMSD}_{\rm tot}$ combines validation and test, excluding RG10, with their existing weights. Panel A transcribes Tables 1 and 3 of Ref.~\citenum{duston2026agentic} (3,547 combined observations). Panels B/C use the local \texttt{agentic2026} reporting variant (870/2,971/1,144 reactions; 3,546 combined): all eight C20C24 reactions move to validation and receive twice their test weight. The paper reports moving two molecules without identifying them; these splits are therefore approximate matches, not identical.
$^{a}$Arithmetic reaggregation of archived cached scores; other local rows use saved full-grid energies. All parameters are frozen. GAS22 uses published coefficients except in the Codex refit. O396, O388, O376, C356, and the Codex refits have no SCF results in this table. Seed-only B97M-V and untested shortlist records cannot be reassigned without their C20C24 test errors. SAFS26-x violates the four checks; SAFS26-a fails the grid check. GAS22 SCF retains the previously disclosed unconverged records.
\end{minipage}
\end{table}

\FloatBarrier

\subsection{GSCDB137: evaluated scope and sensitivity}
Table~\ref{tab:gsc-compact} reports the complete 36-set scope, the 20 sets without MGCDB84 antecedents, and the sensitivity analysis excluding AE11. O159 SCF has mean NER 1.139 versus 1.144 for the local \wb\ reference. Their paired mean difference is approximately $-0.005$ ($p\simeq0.58$; reported bootstrap interval $[-0.055,0.036]$). The two functionals thus give comparable errors on this scope.

\begin{table}[htbp]\centering\small
\caption{Mean GSCDB137 NER on the evaluated subsets (dimensionless).}\label{tab:gsc-compact}
\begin{tabular}{lrrr}\toprule
Functional/mode & All 36 & New 20 & Without AE11 (35)\\\midrule
\wb\ published &1.135&1.285&1.077\\
\wb\ local SCF &1.144&1.290&1.086\\
O159 non-SCF &1.141&1.270&1.101\\
O159 SCF &1.139&1.269&1.099\\
O396 non-SCF &1.147&1.316&1.110\\
GAS22 non-SCF &1.212&1.344&1.227\\
GAS22 SCF &1.190&1.319&1.205\\\bottomrule
\end{tabular}
\par\smallskip\begin{minipage}{0.96\linewidth}\footnotesize
Published GAS22 parameters are used. New denotes sets without an MGCDB84-derived role. Means weight sets equally; no new-functional evaluation covers all 137 datasets. O396 is non-SCF only and C356 has no external evaluation.
\end{minipage}
\end{table}

Including AE11 reduces the GAS22-minus-O159 SCF mean difference from about 0.106 to 0.052. For all 36 sets, the signed-rank $p\simeq0.020$ accompanies a bootstrap mean-difference interval of $[-0.085,0.165]$, which includes zero. The local and published \wb\ rows are distinct because of recorded electronic-state differences; their median absolute molecular-energy difference is approximately $3.6\times10^{-4}$ \kcal, but four molecules differ by more than 1 \kcal.

GSCDB repairs resolved the unconverged classification for O159, although 12 of its 15 repaired records were accepted as energy-stable at a residual gradient floor rather than meeting the strict $10^{-6}$ gradient threshold. GAS22 has 52 such repaired records and two remaining unresolved Pd records. Noise-limited acceptance used a best gradient below $10^{-3}$ and small energy spread among the five smallest-gradient iterates; the largest observed accepted spread was $8.3\times10^{-8}\Eh$. Record-specific tolerances and the density-mask change are preserved with the repair diagnostics. Energy changes were at most $3.9\times10^{-4}$ \kcal\ and mean-NER changes at most $10^{-5}$. These qualifications do not change the reported subset ordering but preclude a claim of uniformly strict convergence.
\FloatBarrier

\section{Non-LLM baseline: SyFES regularized evolution}\label{sec:si-syfes}
\subsection{Implementation and protocol}
The baseline reimplements the SyFES symbolic machinery for the GPU evaluator used in this work. It includes the published instruction pool (five arithmetic instructions with probability 0.06 each, six power instructions with 0.05 each, the $u$-transform with 0.1, and the PBEX, RPBEX, B88X, and PBEC instructions with 0.075 each); the four mutation rules (insertion, removal, replacement, and argument change), each with probability 0.25; a component-selection probability of 1/3 for exchange, same-spin, and opposite-spin factors; at most two bound parameters per enhancement factor; and functional-equivalence fingerprints. GAS22 encoded as instructions reproduces the published closed forms to $10^{-14}$. Evolution settings follow the original study: population 100, tournament size 25, mutation probability 0.9 (otherwise the tournament winner is re-added), and aging removal. Each new form was fitted on training data by CMA-ES with five restarts from $\mathcal N(0,1)$, initial step size 1, bounds $[-10,10]$, function tolerance $10^{-5}$, and at most $10^5$ evaluations per restart. A restart stopped early if all losses were at least $10^5$ or if the best loss exceeded 50 after 5,000 evaluations. Fitness was the validation WRMSD of the restart with the lowest training loss. A child whose fingerprint matched an earlier form re-entered the population with the cached fitness and was not trained. Nonfinite children were discarded. The search stopped after 400 trained forms. The seed was GAS22 with published parameters (float64 training and validation WRMSDs 3.2702 and 3.4623 \kcal).

The implementation differs from the original in six respects that leave the mutation operators and fitness definition unchanged. Only parameters that enter a form are optimized, whereas the original also passed unused shared parameter slots to CMA-ES. Within CMA-ES, nonfinite loss evaluations are assigned $10^{10}$, which reproduces the behavior of the CMA-ES library version used originally. A child already being trained on another worker is skipped. The search runs asynchronously on three GPU workers rather than 50. Fingerprints use the PW\_MOD local-correlation constants of this work. Loss accumulation is adaptive: float32 accumulation (noise about $4\times10^{-5}$ \kcal) is used while the last generation of a restart spans more than 0.01 \kcal, followed by float64 accumulation (noise about $6\times10^{-6}$ \kcal); final training values are recomputed in float64, and the seed's training WRMSD is reproduced as 3.270186 \kcal\ against the float64 reference 3.270184 \kcal. Per-point quantities that do not depend on parameters are precomputed once; each form is compiled into one fused kernel per spin-row type and evaluated for all CMA-ES samples of all five restarts together. After the search, the seed and the ten best forms were re-evaluated in float64 with the evaluator of the main study and screened with the four tests.

\subsection{Results and cost}
The run used 20 h 30 min of wall time on three NVIDIA RTX 6000 Ada GPUs, which were shared with another GPU service throughout, 61.1 GPU-h of fitting (mean 550 GPU-s, median 545 GPU-s, maximum 1,615 GPU-s per form), and $4.77\times10^7$ CMA-ES evaluations (mean $1.19\times10^5$ per form). Of the 400 trained forms, 368 were finite and 51 ended within 60 GPU-s because early stopping removed nonviable mutants. The mutated component was the opposite-spin factor in 156 cases, the same-spin factor in 130, and the exchange factor in 114; the mutation types were 148 insertions, 133 argument changes, 94 replacements, and 25 removals. In addition, 1,079 children were fingerprint-equivalent to earlier forms, 167 tournament winners were re-added without mutation, and four children were skipped as duplicates in training. Among the finite forms, 30 had validation WRMSDs below 3.6 \kcal, 81 below 4.0 \kcal, and 42 above 10 \kcal.

\begin{table}[htbp]\centering\small
\caption{SyFES baseline: float64 re-evaluation of the seed and the ten best forms (stored parameters, \kcal).}\label{tab:si-syfes}
\begin{tabular}{lrrrrl}\toprule
Form & Params & Train & Validation & Violations & Failed test\\\midrule
GAS22 (seed, published) & 14 & 3.2702 & 3.4623 & 1 & uniform-gas exchange\\
SyFES-174 & 14 & 3.4441 & 3.5057 & 0 & ---\\
SyFES-55 & 14 & 3.2690 & 3.5542 & 1 & uniform-gas exchange\\
SyFES-699 & 14 & 3.2072 & 3.5554 & 1 & uniform-gas exchange\\
SyFES-1244 & 14 & 3.1896 & 3.5655 & 1 & uniform-gas exchange\\
SyFES-30 & 14 & 3.1934 & 3.5683 & 1 & uniform-gas exchange\\
SyFES-11 & 14 & 3.1891 & 3.5739 & 1 & uniform-gas exchange\\
SyFES-57 & 14 & 3.1888 & 3.5895 & 1 & uniform-gas exchange\\
SyFES-1266 & 14 & 3.1883 & 3.5944 & 1 & uniform-gas exchange\\
SyFES-784 & 15 & 3.1883 & 3.5945 & 1 & uniform-gas exchange\\
SyFES-542 & 14 & 3.1883 & 3.5949 & 1 & uniform-gas exchange\\
\bottomrule\end{tabular}
\par\smallskip\begin{minipage}{0.97\linewidth}\footnotesize
Form numbers are population identifiers, which include untrained fingerprint-equivalent children. The parameter count of GAS22 includes its two $u$-transform parameters, which the SyFES representation treats as fitted.
\end{minipage}
\end{table}

No trained form improved on the seed's validation error (Table~\ref{tab:si-syfes}). The best-validation trajectory among trained forms reached 3.504 \kcal\ (search value) at the 95th trained form (population identifier 174) and did not improve afterwards. Its training and validation errors are also higher than those of the L-BFGS training refits of the unmodified GAS22 form (Codex: 3.266 and 3.489 \kcal; Opus validation 3.490 \kcal; $u$-transform parameters fixed), the seed used in the agent searches; the added term brings no validation gain. SyFES-174 descends from GAS22 through two fingerprint-equivalent intermediates and is effectively one same-spin mutation: an added $u^2$ term with unit coefficient, where $u=\gamma X/(1+\gamma X)$, followed by a joint refit of all 14 parameters. It passes the uniform-gas exchange test (deviation 0.0032) that GAS22 fails (0.0121). The other nine forms fit training better than GAS22 and validate worse; most add a cubic or sixth-power term or an additional constant to the exchange factor, and all fail the uniform-gas exchange test. The same training--validation tradeoff appears in unconstrained branches of the agent searches.

\subsection{Comparability}
The baseline matches the original study in the evolution algorithm, instruction set, mutation rules, equivalence checking, CMA-ES protocol, fitness definition, data partition, weights, and non-SCF energies on fixed \wb\ densities. Its instructions act only on the scale-invariant features $x^2$ and $w$, as in the original study, so it cannot express an explicit density scale such as the O159 window (the PBEC instruction is applied to $x^2$ and $w$ and introduces no true density scale); the agent searches could additionally use $\rho_\sigma$ (equivalently $r_{s\sigma}$). Quantities such as $\tau_{W\sigma}/\tau_\sigma$ and $\alpha_\sigma$ are functions of $x^2$ and $w$ and were in principle reachable by SyFES, although not by single mutations. The published GAS22 is reproduced at validation 3.462 \kcal\ non-SCF, and on test at 3.599 \kcal\ (non-SCF; 3.585 published) and 3.589 \kcal\ (SCF). The baseline differs from the original study in three respects. First, the original search started from the \wb\ form with all parameters refitted on training data (train/validation/test 2.97/3.82/4.47 \kcal) and evolved to GAS22; the baseline starts from GAS22 itself, the validation-selected minimum of that entire search. It therefore asks whether SyFES can improve on its own published result, not whether it can rediscover GAS22. Second, the original study ran 12 evolutions with 50 workers each and explored 29,628 forms; the baseline is one evolution with about 1,480 proposed children (400 trained and 1,079 equivalent), about 5\% of that exploration. Third, the seed retains its published parameters, whereas each child receives only a training-set CMA-ES fit. CMA-ES applied to the GAS22 form from random starts reaches a training WRMSD of about 3.205 \kcal\ after $2\times10^4$ evaluations per restart and is still decreasing, below the published 3.270 \kcal. The published parameters therefore lie in a different minimum that validates well, and near-miss children may largely be GAS22 refits with an almost inert additional term. The validation error of a CMA-ES refit of the unmodified GAS22 form was not measured; the L-BFGS refit with fixed $u$-transform parameters validates at 3.490 \kcal. The baseline is thus a comparison with the agent searches at a matched number of trained forms on the same split, metric, and non-SCF frame, but with a narrower search space; it is not a replication of the original \wb-to-GAS22 evolution.

\FloatBarrier

\section{Non-SCF token consumption}\label{sec:si-tokens}
\subsection{Definition and aggregate accounting}
At each candidate submission, the evaluation service records the change in cumulative token usage of the run. The total is the sum of uncached input, output, cache-read input, and cache-creation input. Output already includes reasoning-output tokens where reported; these are not added twice. Cumulative run totals sum non-seed charges, including failed attempts. Concurrent proposals, earlier-result summaries, and audit work can be charged to the next submission, so a zero delta does not imply zero model work and a large delta is not necessarily the cost of that record alone.

Table~\ref{tab:tokens-compact} summarizes the runs. Opus contains 397 successful non-seed submissions (396 constraint-clean); the retained Codex run contains 400 successes and one failed attempt. These counts include initialization controls and repeated fits, not only distinct forms. Opus and Codex denote the run archives of the Opus-5.5 and Codex (GPT-6-Astra) agents, respectively.

\begin{table}[htbp]\centering\small
\caption{Assigned non-SCF tokens, in millions except record counts.}\label{tab:tokens-compact}
\setlength{\tabcolsep}{4pt}
\begin{tabular}{lrrrrrr}\toprule
Scope & Records & Input & Output & Cache read & Cache write & Total\\\midrule
Opus &397&0.002&1.500&244.909&5.010&251.421\\
Opus, omit ID 201 &396&0.002&1.285&200.864&3.743&205.893\\
Codex &401&7.852&1.028&512.494&0&521.374\\
Superseded Codex &200&0.829&0.128&31.136&0&32.094\\\bottomrule
\end{tabular}
\par\smallskip\begin{minipage}{0.97\linewidth}\footnotesize
Input is uncached. Values are rounded independently; exact component counts and per-record ledgers are supplied as CSV. The superseded experiment is separate from the 798-attempt retained-run comparison.
\end{minipage}
\end{table}

Opus records 4--200 carry 64,309,206 tokens and records 201--400 carry 187,111,323. ID 201 includes a 45,527,678-token charge booked after the interval between campaigns. It is excluded from the distributions in Figure~\ref{fig:token-distribution} and included in the cumulative progress of Figure~\ref{fig:token-progress}. The Codex total includes 404,586 tokens for failed ID 213. Its final validation-report snapshot contains a further 9,665,130 tokens after the last submission, outside the assigned total of 521,373,961.

Cache reads comprise 97.410\% of the full Opus total and 98.297\% of Codex usage. Mean charges are 633,301 tokens per Opus record (519,931 excluding ID 201) and 1,300,184 per Codex attempt. There are 94 and 17 zero-charge records, respectively. Excluding Opus ID 201, mean output is 3,244.0 tokens per record versus 2,562.6 for Codex. Thus, the larger Codex total predominantly reflects context/input accounting, not a proportionate increase in generated text.

\subsection{Iteration profiles and validation progress}
\begin{figure}[p]\centering
\includegraphics[width=0.93\linewidth,height=0.72\textheight,keepaspectratio]{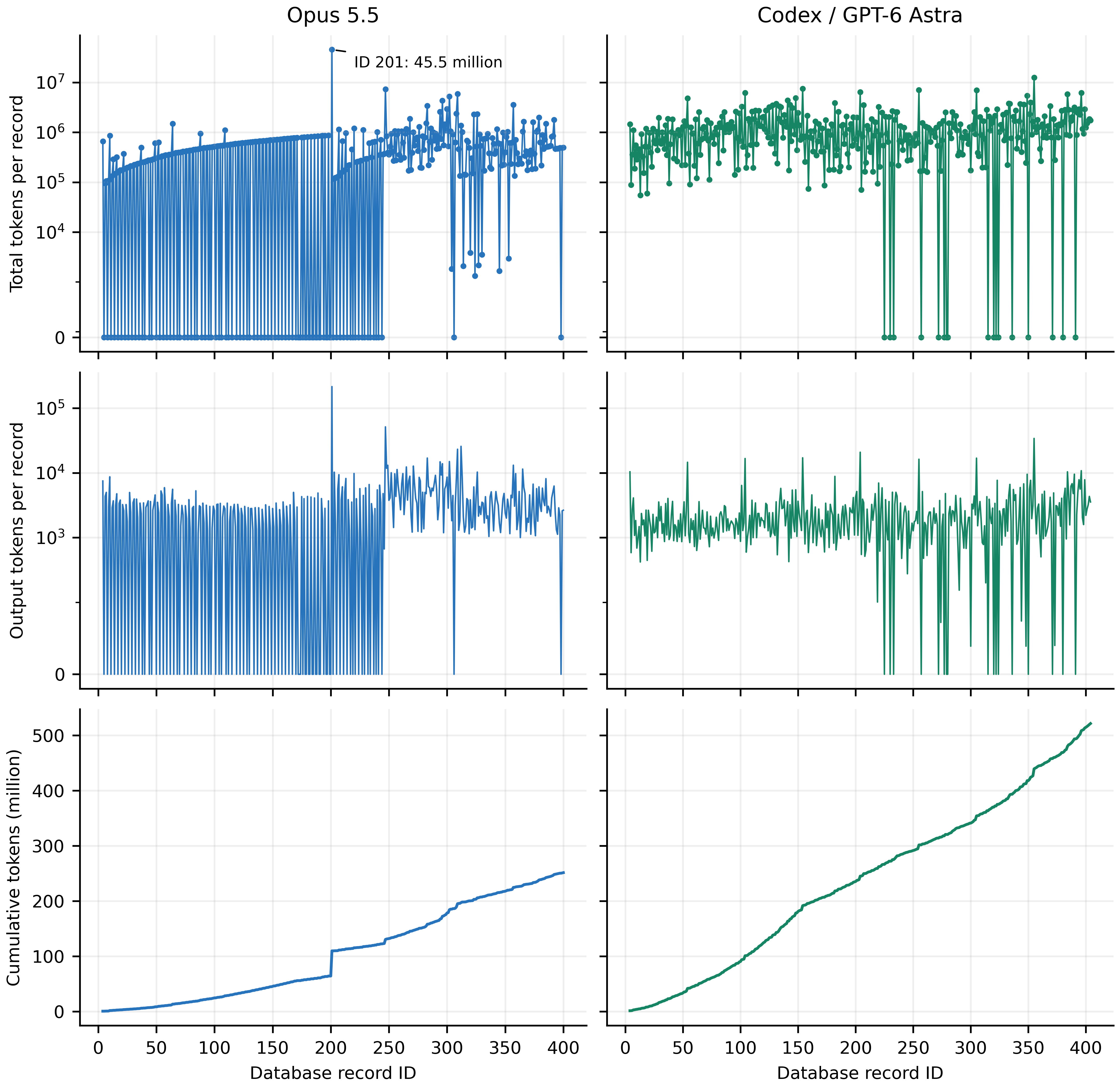}
\caption{Token charges by database record ID for the retained runs: 397 Opus and 401 Codex attempts. Rows show total, output, and cumulative charges. The first two use symmetric-log axes that retain zeros; cumulative axes are linear. The Opus ID 201 inter-campaign charge and Codex failure 213 are included. Input and cache components are available for every record in the companion CSV files.}\label{fig:token-iterations}
\end{figure}

\begin{figure}[htbp]\centering
\includegraphics[width=\linewidth]{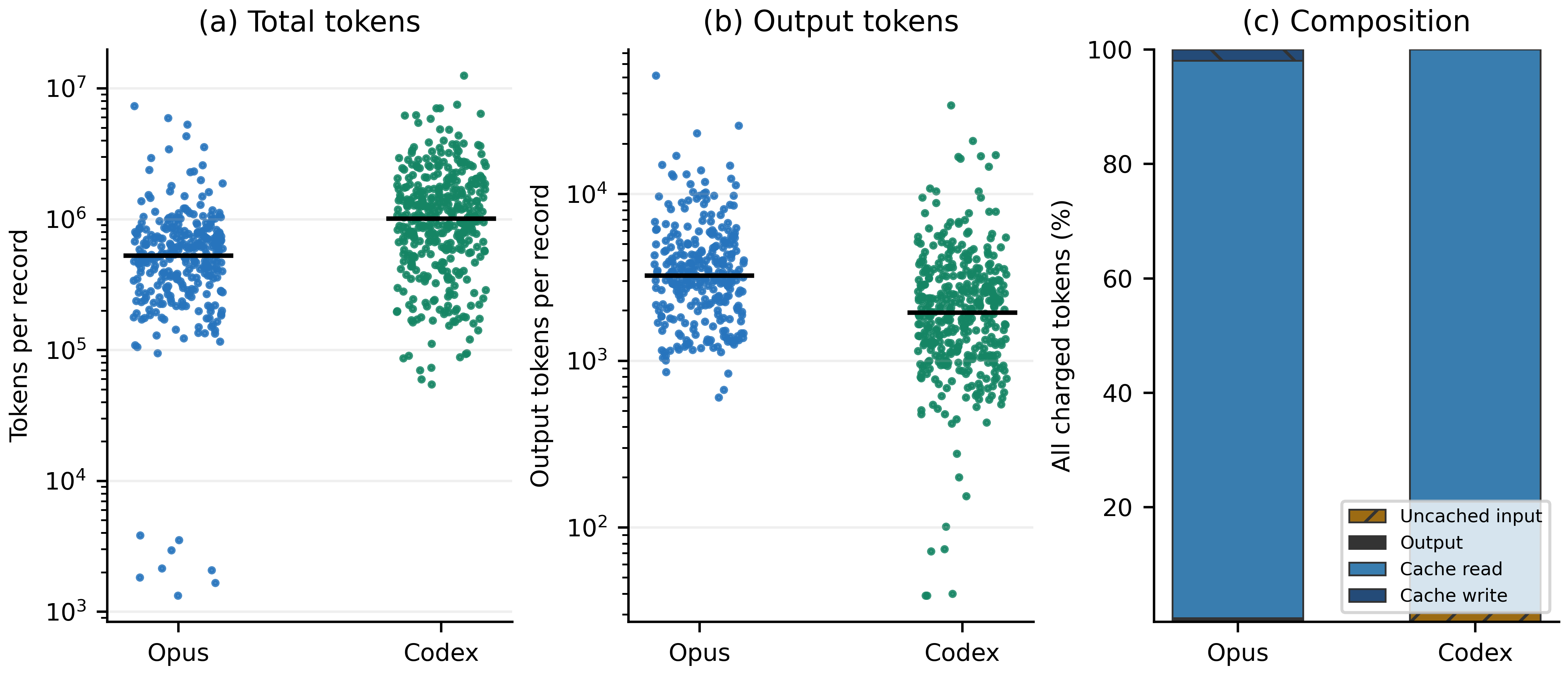}
\caption{Token distributions and composition. Distribution panels (a,b) omit Opus ID 201 and omit zero charges on logarithmic axes; bars indicate nonzero medians. Composition panel (c) includes all charges, including ID 201. The arithmetic means in the text include zero-charge records within each stated scope.}\label{fig:token-distribution}
\end{figure}

\begin{figure}[htbp]\centering
\includegraphics[width=0.93\linewidth]{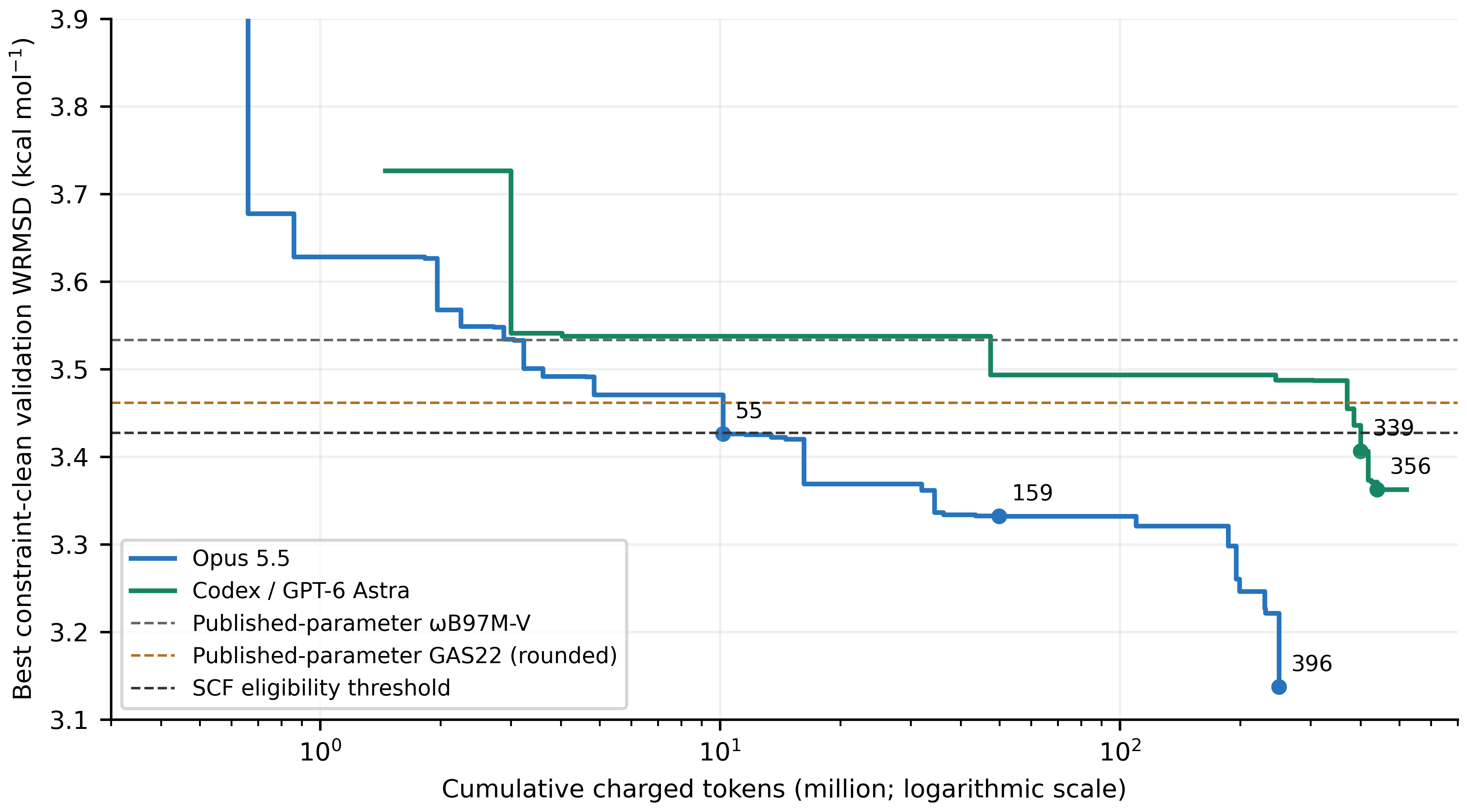}
\caption{Cumulative assigned tokens versus best constraint-clean validation WRMSD. All charges enter the horizontal axis, including Opus ID 201 and Codex failure 213. Values above 3.9 \kcal\ are outside the displayed range but remain in the archive. Passing evaluator checks is distinct from passing later source/parsimony audits. No held-out test or external metric updates the curves.}\label{fig:token-progress}
\end{figure}

Figures~\ref{fig:token-iterations}--\ref{fig:token-progress} summarize token use by iteration. Exact per-iteration components and cumulative values are provided in two run-specific CSV files in the reproducibility archive.

The first constraint-clean SCF-gate crossings occur at O55 after 10,169,271 assigned tokens and C339 after 399,861,159. O159 is reached at 49,926,626 tokens, O396 at 249,942,965, and C356 at 440,604,859. O145 to O159 adds only a 0.000562 \kcal\ validation improvement over an interval carrying 6,409,234 tokens across multiple proposals. Codex refinement from C339 through C350 to C356 lowers validation from 3.406690 to 3.373555 to 3.362805 while worsening noncovalent-dimer errors. Its eventual test outcome reverses the validation advantage over the refitted GAS22 reference.

The retained searches stopped at their assigned iteration/evaluation budgets, not at a demonstrated global optimum. Opus ends at ID 400; Codex stops after 400 successes plus one failed attempt. No later Codex record improves C356's raw validation score. The summed evaluator durations are approximately 11.09 and 8.56 h, distinct from elapsed calendar time and token usage. The Codex run was interrupted and resumed within the same experiment, so its elapsed time exceeds its computation time.

The superseded 200-attempt Codex experiment had 198 successes and two infrastructure failures, with best validation 3.495181443 at ID 107, above the SCF gate. Its verified assigned total is 32,093,724 tokens, but only 44 per-record totals and 39 full component splits were recoverable. Missing entries are explicitly blank in its separate CSV; no interpolation is used.
\FloatBarrier

\section{Agent workflow and representative exchanges}\label{sec:si-workflow}
The multi-agent framework combines a coordinator, thematic islands, shared lineage/memory, and numerical evaluator workers through MCP.\cite{hou2026model} Each iteration records a proposal before evaluation and an outcome summary afterward, following the Plan--Execute--Summarize organization used in SAFS26.\cite{duston2026agentic} The completed Codex run used four islands, audits every 50 successes, and at most three concurrent GPU evaluations. Final successful allocations were 100/120/100/80 after a documented budget transfer. The Opus archive contains thematic lineages A--H across two continued campaigns.

The Codex shortlist used validation eligibility followed by a 0.02 \kcal\ tolerance band favoring fewer parameters, a smaller validation--training gap, and fewer category regressions. Source and parameter hashes were frozen before test access. Only C356 and its two authorized refitted references were subsequently tested. Audit exclusions remained distinct from numerical constraint failures: the final Codex review excluded 29 records, and 17 Opus continuation records were excluded for frozen published exchange coefficients.

Persistent plans and summaries make these distinctions auditable. The O145-to-O159 record separates the proposed density-window change from its very small validation gain, while the C339-to-C356 sequence records both aggregate improvement and worsening noncovalent categories. Source audits exposed unused parameters, intended changes absent from code, and frozen-coefficient variants. The records therefore document the search decisions themselves. Because the agents proposed and logged refitted ablations as part of the search, the same records also supplied every control in Table~\ref{tab:si-window-controls}; the window analysis of Section~\ref{sec:si-window} required no additional fitting of the controls.

\subsection{Generic initial prompt}
The box below is a representative condensation of the task instructions; the list of example directions is quoted from the original prompt. The continuation prompt for the second Opus campaign assigned island themes that built on the first campaign, including refinement of the same-spin window. Original prompts are preserved in the archive.
\begin{taskbox}{Task instructions --- representative reconstruction}
Coordinate a multi-agent search for compact exchange-correlation functional forms. Keep omega = 0.3 bohr^-1, short-/long-range HF fractions = 0.15/1.0, and VV10 b = 6, C = 0.01 fixed.

Modify the semilocal exchange and same-/opposite-spin correlation factors using spin density, gradients, and positive kinetic-energy density. Explore structurally different ideas, for example new bounded descriptors, rational/Pade forms with positive denominators, r_s- or zeta-dependent correlation, iso-orbital (alpha) regimes, different gradient/kinetic transforms, and correlation forms that obey known limits (e.g. one-electron self-correlation freedom via tau_W/tau, high-density scaling). Submit one Functional subclass with all fitted coefficients and empirical shape parameters in init_params(). Do not freeze fitted parent coefficients as hidden constants.

Fit on 870 training reactions; use 2963 validation reactions to compare forms. The 1152 test reactions remain unavailable during search. Request at most 3 restarts and 500 optimizer iterations; select restarts by training loss only.

Before evaluation, record the algebraic change, initialization, parameter count, rationale, and constraint risks. Inspect aggregate/category errors, spin symmetry, uniform-gas exchange, scaling, grid sensitivity, and numerical finiteness. Afterward, record the actual outcome, including failures and tradeoffs.

Use shared lineage and memory, thematic islands, periodic migration, and independent source/parameter audits. At the assigned budget, freeze the shortlist. Test, SCF, and external evaluation are separate reporting stages, not feedback for revising this search.
\end{taskbox}

\subsection{Density-window refinement: O145 to O159}
These exchanges are abridged reconstructions from submitted plans, source changes, and saved outputs, not verbatim transcripts.
\begin{proposalbox}{Proposal --- reconstructed from the O159 record}
Replace the parent's r/(1+r^3) same-spin window by r/(1+r^4), aiming to reduce the correction in diffuse regions while retaining an intermediate-density effect. Keep 13 parameters, initialize from O145, and jointly refit all retained coefficients.
\end{proposalbox}
\begin{feedbackbox}{Evaluator feedback --- rounded saved values, kcal/mol}
                 O145          O159
Training         3.225679      3.251690
Validation       3.332936      3.332374
Violations       0             0
O159 charge: 733348 total tokens; 3550 output tokens.
\end{feedbackbox}
\begin{modelbox}{Outcome summary --- evidence-based reconstruction}
The cutoffs have nearly equal measured validation errors. O159 has the smaller training-validation gap, with mixed category changes. The result does not establish a unique physical preference for exponent four. Later SCF/external results were not available in this evaluator response.
\end{modelbox}

\subsection{Screen refinement and coefficient tying: C339 to C356}
\begin{proposalbox}{Successive proposals --- reconstructed from source and plans}
C350: replace 1/(1+y) by 1/sqrt(1+y^2), where y = softplus(q)*r_s, in the opposite-spin kinetic correction. Keep 15 parameters and refit.

C356: tie the pure kinetic pair as a*W^2*(1-W^4). Retain independent root couplings, the same-spin orbital window, and the density screen. Reduce to 14 jointly fitted parameters.
\end{proposalbox}
\begin{feedbackbox}{Evaluator feedback --- rounded saved values, kcal/mol}
                 C339          C350          C356
Training         3.226480      3.189786      3.207428
Validation       3.406690      3.373555      3.362805
Violations       0             0             0
C356 NCED/NCD regression vs GAS22 refit: 10.85% / 5.61%.
\end{feedbackbox}
\begin{modelbox}{Outcome summary --- evidence-based reconstruction}
Aggregate validation improves with one fewer coefficient, but category regressions remain. Joint refitting does not isolate the coefficient tie as the unique cause. After the search was frozen, C356's test error was 3.897343 versus 3.627617 for its GAS22 refit; this later outcome was not search-time feedback.
\end{modelbox}

\section{Data and reproducibility}
The source code is available at \url{https://github.com/BaratiLab/DFT_functional_agentic_searcher}. The accompanying archive supplies full-precision parameters and functional source, exact token ledgers, all split/category/set metrics, 5,930 MGCDB84 and 3,487 GSCDB137 absolute-energy records with convergence flags, benchmark geometry/input files, and figure-generation code. Its README maps all manuscript and SI figures to their data and scripts. Exhaustive numerical tables and precursor formulas are provided in the archive.

Arithmetic verification reproduced 40 full-grid MGCDB84 aggregates and 1,080 GSCDB137 metric cells from the archived energies, with maximum discrepancies of $8.9\times10^{-16}$ \kcal\ and zero in the recorded Python 3.12 environment. All 9,417 geometry/input hashes were checked. Independent transcriptions of O159, O338, O396, and C356 matched their source on 2,048 synthetic spin-polarized points each. These checks establish data consistency and equation transcription. The attenuation-branch correction is quantified separately in Section~\ref{sec:si-attenuation}; its reevaluation scripts and per-molecule shifts for all 5,930 MGCDB84 and 3,487 GSCDB137 calculations are included in the archive, as are the window-analysis scripts, descriptor bins, recorded predictions, and per-molecule O178 semilocal differences for the GSCDB137 control. Large reference-density and feature caches are not included; the scripts that regenerate them are.

\FloatBarrier

\clearpage
\bibliography{references}

\end{document}